\documentclass[10pt,journal,compsoc]{IEEEtran}

\usepackage[pdftex]{graphicx}
\usepackage{amsmath}
\usepackage{amssymb}
\usepackage{amsthm}
\usepackage{bm}
\usepackage{enumerate}
\usepackage{balance}
\usepackage{hhline}
\usepackage{array}
\usepackage{url}
\usepackage{xspace}
\usepackage{wasysym}
\usepackage{booktabs}
\usepackage{multirow}
\usepackage{setspace}
\usepackage{pifont}
\usepackage{dsfont} 
\usepackage[inline]{enumitem}
\usepackage{ctable}
\usepackage{color,colortbl}

\usepackage[ruled,vlined,linesnumbered]{algorithm2e}
\usepackage[symbol]{footmisc}
\usepackage{multirow}

\usepackage{xcolor}

\usepackage[pagebackref=true,breaklinks=true,colorlinks,bookmarks=false]{hyperref}
\usepackage[capitalize]{cleveref}

\ifCLASSOPTIONcompsoc
  \usepackage[nocompress]{cite}
\else
  \usepackage{cite}
\fi

\definecolor{Gray}{gray}{0.9}
\definecolor{White}{gray}{1}
\definecolor{DGray}{gray}{0.8}
\definecolor{DDDDGray}{gray}{0.3}
\definecolor{citecolor}{HTML}{0071bc}

\definecolor{DeltaColor}{rgb}{0.039,0.73,0.71}
\definecolor{SigmaColor}{rgb}{0.98,0.45,0.0}
\definecolor{AlphaColor}{rgb}{0,0,0.8}
\definecolor{BetaColor}{rgb}{0.8,0,0.8}
\definecolor{GammaColor}{rgb}{0.514,0.34,0.224}
\definecolor{EpsilonColor}{rgb}{0.353,0.725,0.906}
\definecolor{GreenColor}{rgb}{0.137,0.573,0.565}
\definecolor{RedColor}{rgb}{0.949,0.275, 0.224}
\hypersetup{linkcolor=black, citecolor=black, urlcolor=black}

\newcommand\ie{\textit{i.e.}\xspace}
\newcommand\eg{\textit{e.g.}\xspace}
\newcommand\etal{\textit{et al.}\xspace}
\newcommand\vs{\textit{vs.}\xspace}
\crefname{table}{Tab.}{Tabs.}

\newcommand\emath{\ensuremath}
\newcommand\world{$\mathcal{W}$\xspace}

\newcommand\Query{\texttt{\textbf{Query}}\xspace}
\newcommand\Basis{\texttt{\textbf{Basis}}\xspace}
\newcommand\POEMs{POEM$_{\text{v1}}$\xspace}
\newcommand\POEMg{POEM$_{\text{v2}}$\xspace}

\newcommand\model{POEM\xspace}
\newcommand\dexycbmv{DexYCB-Mv\xspace}
\newcommand\hodddmv{HO3D-Mv\xspace}
\newcommand\oakinkmv{OakInk-Mv\xspace}
\newcommand\interhandmv{InterHand-Mv\xspace}
\newcommand\arcticmv{Arctic-Mv\xspace}

\newcommand\MPVPE{MPVPE$\downarrow$}
\newcommand\RRV{RR$_{\rm{V}}$$\downarrow$}
\newcommand\PAV{PA$_{\rm{V}}$$\downarrow$}
\newcommand\AUCVtwenty{AUC$_{\rm{V}@20}$$\uparrow$}
\newcommand\AUCVfifty{AUC$_{\rm{V}@50}$$\uparrow$}
\newcommand\MPJPE{MPJPE$\downarrow$}
\newcommand\RRJ{RR$_{\rm{J}}$$\downarrow$}
\newcommand\PAJ{PA$_{\rm{J}}$$\downarrow$}
\newcommand\AUCJtwenty{AUC$_{\rm{J}@20}$$\uparrow$}
\newcommand\AUCJfifty{AUC$_{\rm{J}@50}$$\uparrow$}
\newcommand\AUCJ{AUC$_{\rm{J}}$$\uparrow$}
\newcommand\AUCV{AUC$_{\rm{V}}$$\uparrow$}

\newcommand{\del}[1]{\xspace}

\newcommand{\paraheading}[1]{\vspace{5pt}\mbox{\textbf{#1}\;}}

\DeclareMathAlphabet\mathbfcal{OMS}{cmsy}{b}{n}

\newlength\savewidth

\begin{document}
%
\title{Multi-view Hand Reconstruction with a Point-Embedded Transformer}
%
%
%
%

\author{Lixin Yang,~
  Licheng Zhong,~
  Pengxiang Zhu,~
  Xinyu Zhan,~
  Junxiao Kong,~
  Jian Xu,~\\
  and~Cewu Lu,~\IEEEmembership{Member,~IEEE}
  \IEEEcompsocitemizethanks{
    \IEEEcompsocthanksitem Lixin Yang and Cewu Lu are with the School of Artificial Intelligence (SAI), Shanghai Jiao Tong University, Shanghai 200230, China (e-mail: siriusyang@sjtu.edu.cn; lucewu@sjtu.edu.cn). Cewu Lu is the corresponding author.
    \IEEEcompsocthanksitem Licheng Zhong is with the School of Mechanical Engineering, Shanghai Jiao Tong University, Shanghai 200240, China (e-mail: zlicheng@sjtu.edu.cn).
    \IEEEcompsocthanksitem  Pengxiang Zhu, Xinyu Zhan, and Junxiao Kong are with the School of Electronic Information and Electrical Engineering, Shanghai Jiao Tong University, Shanghai 200240, China (e-mail: zhu\_peng\_xiang@sjtu.edu.cn; kelvin34501@sjtu.edu.cn; alive-x@sjtu.edu.cn)
    \IEEEcompsocthanksitem Jian Xu is with the Institute of Automation Chinese Academy of Sciences (CASIA), Beijing 100190, China (e-mail: jian.xu@ia.ac.cn).
  }
}

%
%

\markboth{IEEE TRANSACTIONS ON PATTERN ANALYSIS AND MACHINE INTELLIGENCE, VOL. 47, NO. 11, NOVEMBER 2025}%
{****for peer review only****}
%



\IEEEtitleabstractindextext{%
  \begin{abstract}

    This work introduces a novel and generalizable multi-view Hand Mesh Reconstruction (HMR) model, named POEM, designed for practical use in real-world hand motion capture scenarios.
    The advances of the POEM model consist of two main aspects.
    First, concerning the modeling of the problem, we propose embedding a static basis point within the multi-view stereo space. A point represents a natural form of 3D information and serves as an ideal medium for fusing features across different views, given its varied projections across these views. Consequently, our method harnesses a simple yet effective idea: a complex 3D hand mesh can be represented by a set of 3D basis points that 1) are embedded in the multi-view stereo, 2) carry features from the multi-view images, and 3) encompass the hand in it.
    The second advance lies in the training strategy. We utilize a combination of five large-scale multi-view datasets and employ randomization in the number, order, and poses of the cameras. By processing such a vast amount of data and a diverse array of camera configurations, our model demonstrates notable generalizability in the real-world applications.
    As a result, POEM presents a highly practical, plug-and-play solution that enables user-friendly, cost-effective multi-view motion capture for both left and right hands. The model and source codes are available at \url{https://github.com/JubSteven/POEM-v2}.
  \end{abstract}

  }

\maketitle

\IEEEdisplaynontitleabstractindextext

\ifCLASSOPTIONpeerreview
  \begin{center} \bfseries EDICS Category: 3-BBND \end{center}
\fi
%
\IEEEpeerreviewmaketitle


%
%
%

\IEEEraisesectionheading{\section{Introduction}\label{sec:introduction}}

%
%
%
%

\IEEEPARstart{Hand}{}
plays a central role in human activity, as it is the primary tool used to interact with the 3D world.
As Anaxagoras noted, ``\textit{it is because of being armed with hands that humans are the most intelligent animals}''.
Therefore, understanding human hand motion is pivotal for researchers aiming to equip intelligent robots with human-level skills.
For example, vision-based teleoperation requires tracking hand motion from human demonstrations to effectively map these motions to robot hand.
This process necessitates accurately recovering human hand's geometry (pose and surface).
Recently, significant efforts have been made in monocular 3D Hand Mesh Reconstruction (HMR) \cite{Park2022HandOccNet,Pavlakos2024hamer,chen2025handos,dong2024hamba}.
However, challenges still remain in producing applicable results, mainly due to the following reasons:
\begin{enumerate*}[label={\textbf{\roman*)}}]
    \item \textbf{Depth ambiguity.} Recovering the absolute position in a monocular camera system is an ill-posed problem. Consequently, previous methods \cite{ge20193d,Wang2018Pixel2Mesh} only recovered hand vertices relative to the wrist (\ie, root-relative).
    \item \textbf{Unknown perspective.} The position of the hand’s 2D projection is highly dependent on the camera’s perspective model (\ie camera intrinsics). Monocular methods commonly suggest a weak perspective projection \cite{lin2021metro,Pavlakos2024hamer}. As a result, hands reconstructed using these methods possess only an intuitive depth.
    \item \textbf{Occlusion.} Occlusion between the hand and interacting objects poses a challenge to the accuracy of reconstruction, particularly evident in monocular setup \cite{Park2022HandOccNet}.
\end{enumerate*}
These issues limit the practical application of monocular-based methods, especially in scenarios requiring interaction with surroundings, where the absolute and accurate position of the hand model is crucial.

This paper thus focuses on building a generalizable model for HMR under multi-view setting and develops a cost-effective, user-friendly multi-view hand capturing system for seamless deployment to every user.
Our motivation stems from three aspects:
First, the three issues mentioned above can be addressed by leveraging the spatial consistency among multi-view geometry.
Second, the availability and continuous expansion of large-scale multi-view hand datasets make it possible to develop a \textbf{generalizable pretraining multi-view HMR model}.
Then, our model, through providing an off-the-shelf hand reconstruction pipeline, can offers an appealing solution for vision-based teleoperation.

In the multi-view setting, we focus on two main parts: 1) aligning observations from different camera spaces into a common representation space and fusing them, and 2) designing network architectures in this representation space to make predictions. \textbf{1) Alignment.} Since multi-view setups vary across datasets, platforms, and applications,
we seek a basis function that is invariant to camera configurations to facilitate this alignment process.
We draw inspiration from the Fourier Series, wherein any periodic function can be represented by a series of sinusoidal functions (bases). Using this analogy, we propose a set of static 3D points in world space as the basis for representing the variable hand surface points.
These 3D basis points lie within the common view space of all cameras and encompass the hand for reconstruction.
The alignment proceeds as follows.
The camera extrinsic are applied directly to the basis points in world space to transform them into each camera space.
In this way, each basis point will be projected onto different pixel positions of different cameras, and the corresponding image feature is sampled at these positions.
The basis points, now carrying $N$ sampled features from $N$ camera views, are brought back to world space to represent the multi-view observations in a unified space.
Then, We design a point-wise feature fusion module, Projective Aggregation, This module fuses the $N$ camera features for each basis point via a non-local network.

\textbf{2) Prediction.}
As the basis points themselves processes spatial relations, and the point-wise fused features retain the camera configurations, we propose a spatial-aware network to process these basis points and make predictions for the hand.
We design a Point-Embedded Transformer with a spatial-aware point cloud attention in it. This Transformer takes two point clouds as inputs: the basis points and the initial hand points, where the latter are initialized using a hand template, and progressively transforms the features on the basis points to the hand points and updates the coordinates of the hand points.

We name this model \textbf{\model}, highlighting its two key components: the \textbf{Po}int-\textbf{Em}bedded common-view space and the \textbf{Po}int-\textbf{Em}bedded Transformer.
Using basis points as the media for feature fusing and network processing avoids encoding camera extrinsics into the network, thereby enabling the scaling-up of training data from different datasets.
Recent progress in monocular HMR \cite{Pavlakos2024hamer} verifies that increasing the scale of training data is the key to a generalizable foundation model.
To ensure that our model can generalize across any multi-view configuration, we use a mixture of five large-scale multi-view hand datasets for training. For each multi-view sample, we randomly reduce the number of cameras and shuffle their order to further enhance the dataset scale.
Our model successfully consume such large amount of training data and consistently demonstrate outstanding performance across all test sets.
Additionally, by comparing our alignment and prediction modules with other state-of-the-art models, our model consistently outperforms alternative designs.

\model has been deployed on a real-world multi-view camera platform, paring with a  high-precision commercial motion caption system to obtain ground-truth hand points.
This platform records the user's bimanual object manipulation tasks on a desktop.
Subsequently, we compare the \model's predictions with another method that employs iterative optimization on 2D keypoint estimation \cite{zhang2020mediapipe}. POEM model shows marked advantages in both performance and speed. We believe that this generalizable multi-view hand reconstruction model can be deployed in any real-world scenario. With just a few cost-effective cameras, it facilitates the motion capture for manipulation on a global scale.
\section{Related Work}
\paraheading{Multi-view Feature Processing.}
Representing the observations from different camera systems in a unified way while fusing multi-view features accordingly is a common problem in multi-view stereo (MVS) reconstruction and pose estimation. This literature review focuses on addressing this key challenge.
From this perspective, previous methods - in which the camera transformation (extrinsic) is typically encoded differently - can be seen as different types of Position Embedding (PE).
For example, the method based on epipolar transform \cite{He2020EpipolarT} can be classified as a line-formed PE, as pixels in one camera are encoded as epipolar lines in others.
Additionally, there are point-formed position embeddings such as FTL in \cite{Remelli2020LightweightM3, Han2022UmeTrack}, RayConv in \cite{Wang2021MVP}, and 3D Position Encoder in PETR \cite{Liu2022PETR}, which apply camera extrinsic directly to point-shaped features (FTL) or add camera ray vectors channel-wise to the features (RayConv, PETR).
These two point-formed PEs use point-formed features solely in 2D format because 2D convolution or image-based self-attention cannot capture 3D structure.
Therefore they are considered implicit.
In contrast, 3D CNN and point cloud network can preserve 3D structure. SurfaceNet \cite{ji2017surfacenet} and LSM \cite{kar2017LSM} associate features from different views by forming a cost volume and rely on 3D CNNs to perform voxel-wise reconstruction. To address the drawback of the final volumetric output, MVSNet \cite{yao2018mvsnet} predicts the depth-map instead of voxels. These methods are classified as voxel-formed PEs.
Finally, the explicit point-formed PE directly uses a set of 3D points in the MVS scene.
For instance, PointMVS \cite{chen2019PointMVSNet} unprojects the predicted depth-map to a point cloud and aggregates features from different views using project-and-fetch. Our method also belongs to this type.
In our task, preserving the topology of the hand vertex points is crucial, but this can be challenging with PointMVS, which indiscriminately treats points. Instead, our approach, \model, represents the common-view scene as an unstructured point cloud for feature aggregation and employs a structure-aware vertex query to initialize and update hand vertices. \model's structured vertices interact with unstructured frustum points through cross-attention, which effectively removes object occlusion and accurately reconstructs the hand mesh.

\paraheading{Monocular Hand Reconstruction.}
Monocular hand reconstruction has been a long-studied topic. A series of works \cite{hasson2019learning, Pavlakos2024hamer, Kong2022IdentityAwareHM} were built upon the deformable hand mesh with a differentiable skinning function, \textit{e}.\textit{g}.\  MANO \cite{romero2017embodied}.
However, the difficulty of regressing the non-Euclidean rotations hinders the performance of these methods.
There have been emerging works exploring the direct reconstruction of the hand surface. As vertices naturally lie in 3D Euclidean space,
\cite{ge20193d,chen2021camera} leveraged the mesh structure of MANO with the graph-based convolution networks (GCN). Besides, the voxels \cite{moon2020i2l}, UV positional maps \cite{chen2021i2uv}, and signed distance function \cite{karunratanakul2020grasping} were also competent choices. Recently, Transformers \cite{lin2021metro,lin2021graphormer} have been deployed to fuse the features on the hand surface points by self-attention mechanism.
In this work, we follow the path of direct mesh reconstruction with Transformer and propose to model hand mesh as a \textbf{point set} in the multi-view stereo.

\paraheading{Point Cloud Network.}
Qi \etal \cite{qi2017pointnet} proposed PointNet, the first deep model utilizing the permutation-invariant structure of point cloud data. A subsequent work,  PointNet++, was proposed in \cite{qi2017pointnet++} thereafter. PointNet++ introduced ball query and hierarchical grouping, enabling the model to reason form local structures of point clouds. There were a number of successive works \cite{li2018pointcnn,thomas2019kpconv} trying to define local convolution operators on point clouds to extract local information.
The recent application of transformers on point cloud data has been proven a success. Zhao \textit{et al}.\  proposed Point Transformer \cite{Zhao2021PointT}, which adopts a vector attention mechanism to perform attention in the local neighborhood. Guo \textit{et al}.\  proposed the Point Cloud Transformer \cite{guo2021pct}, which used an analogy of the Laplacian matrix on point clouds to fuse long-range relationships in the point cloud.
Our method follows the design in Point Transformer, but selectively fuses the features on basis points to a hand vertex through a \textbf{cross-set} vector attention.

%
%
%
%

\begin{figure*}[!t]
    \centering
    \includegraphics[width=0.9\linewidth]{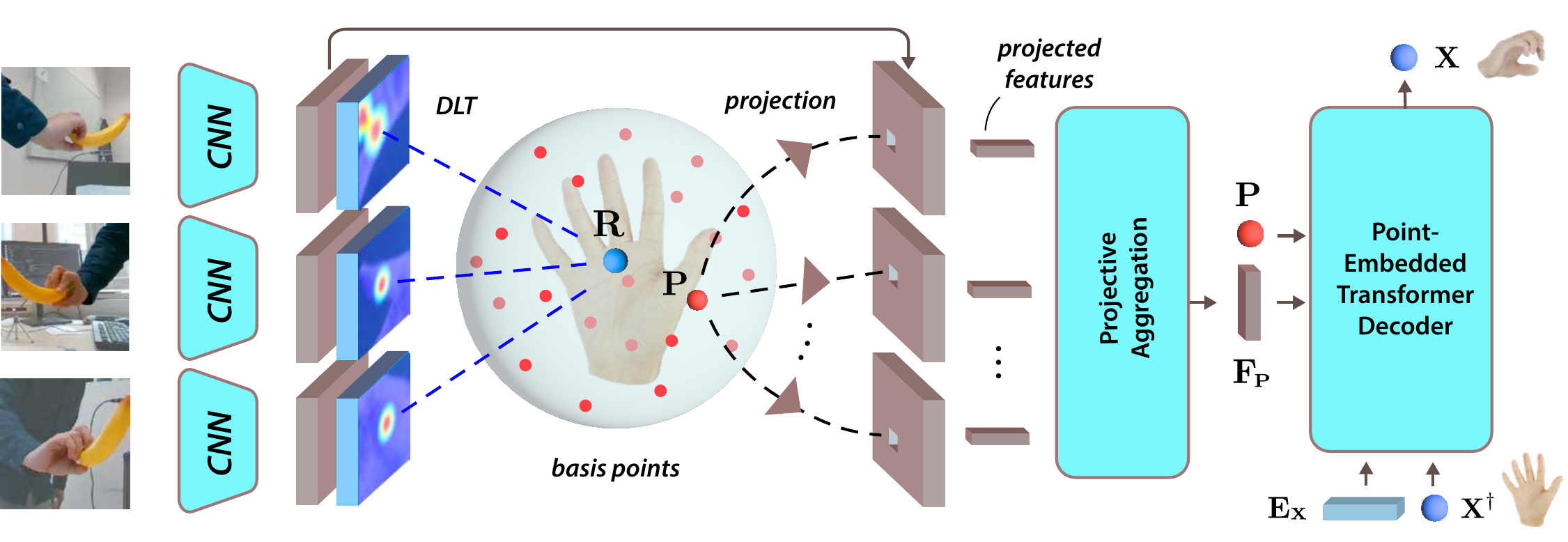}
    \vspace{-3mm}
    \caption{The architecture of \model model. The first stage estimates the root position $\mathbf{R}$ and the second stage reconstructs the query points $\mathbf{X}$.}
    \label{fig:model_architecture}
\end{figure*}
\section{Formulation}\label{sec:formulation}

The general purpose of POEM is to model the joint distribution of hand skeleton joints and surface vertices under multi-view observations.
Given \textit{N} cameras with different positions and orientations, we define the 3D coordinates of the joints and vertices inside a shared 3D world space: \world.
Inside this space,  $\mathbf{J} \in \mathbb{R}^{21\times3}$ represents the 3D joints on the hand skeleton, and $\mathbf{V} \in \mathbb{R}^{778\times3}$ denotes the 3D vertices on the hand surface mesh,
Given a image set $\mathcal{I} = \{\mathbf{I}_v\}^{N}_{v=1}$ from total \textit{N} views, the proposed POEM model learns the distribution:
$P(\mathbf{J}, \mathbf{V} ~|~ \mathcal{I})$.
As both the joints and vertices are 3D points in \world, we subsequently refer to them collectively as the \textbf{Query Points}, denoted as $\mathbf{X} = [\mathbf{V}, \mathbf{J}] \in \mathbb{R}^{799\times3}$.

Learning the absolute positions of these query points in an unconstrained 3D space proves to be challenging.
Since the 3D position of \emath{\mathbf{X}} in \world can be interpreted as the sum of a root point \emath{\mathbf{R} \in \mathbb{R}^{3}} situated in \world and the root-relative query points \emath{\mathbf{X}_{rel} \in \mathbb{R}^{799\times3}} that are offset with respect to the root point, as:
\begin{equation}
    \mathbf{X} = \mathbf{X}_{rel} + \mathbf{R},
    \label{eq:root_rel}
\end{equation}
we can decompose the joint distribution based on the chain rule as follows:
\begin{equation}
    \begin{aligned}
        P(\mathbf{X} ~|~ \mathcal{I}) & = P(\mathbf{X}_{rel}, \mathbf{R} ~|~ \mathcal{I})                                                                  \\
                                      & = P_{\bm{\omega}}(\mathbf{X}_{rel} ~|~ \mathbf{R}, \mathcal{I}) \cdot P_{\bm{\tau}}(\mathbf{R} ~|~ \mathbfcal{I}).
        \label{eq:formulation}
    \end{aligned}
\end{equation}
The training process is to fit the learnable parameters $\bm{\tau}, \bm{\omega}$ on the training data, \ie
for each input $\mathcal{I}$, the model maximizes its probability at the ground-truth: \emath{\widehat{\mathbf{X}}_{rel}} and  \emath{\widehat{\mathbf{R}}}.
This formulation expresses \model's two-stage architecture (shown in \cref{fig:model_architecture}). In the following sections, we will first introduce the approach to determining the absolute root position, \emath{\mathbf{R}}, in \cref{sec:root_estim}, and then elaborate on the method for obtaining the root-relative query points, \emath{\mathbf{X}_{rel}}, in \cref{sec:query_estim}.

\section{Root Point from Triangulation}\label{sec:root_estim}

The first stage of \model is to predict the 3D root point $\mathbf{R}$ from the multi-view images $\mathbfcal{I}$.
To achieve this,
we first estimate the $\mathbf{R}$'s 2D pixel-level coordinates in each image view $\mathbf{I}_i$ and then lift them to the 3D space through algebraic triangulation.

\subsection{Heatmap-based Root Estimation}
This first stage is achieved by a Convolution Neural Network (CNN) that has pretrained on a large-scale image dataset (\ie ImageNet).
The CNN backbone performs progressively feature extraction via a series of residual blocks.
In each of the hidden blocks, the resolution of feature maps is reduced by a factor of 2, and the number of channels is increased.
To leverage the features from different scales, we add a series of deconvolution layers
that recursively upsample the feature map from the current layer and concatenate it with the feature map from the preceding layer.
This process continues until the feature map reaches the $1/8$ of the original image resolution.
The final layer is a $1\times1$ convolution layer that maps this multi-scale image feature into a 2D likelihood heatmap.
A learned heatmap $\mathbf{H}$ assigns a probability for each location, indicating the likelihood of the root point being at that location.
Accordingly, we apply a soft-argmax operation to calculate the root's 2D coordinates $\mathbf{r}$ (expectation of the heatmap) as follows:
\begin{equation}
    \mathbf{r} = \sum_{v}\sum_{u} (u, v) \cdot \mathbf{\tilde{H}}(u,v),
    \label{eq:softargmax}
\end{equation}
where the $u, v$ are the 2D feature coordinate in the horizontal and vertical directions, respectively, and $\mathbf{\tilde{H}}$ is the normalized heatmap.
The normalization process is defined as: $ \mathbf{\tilde{H}} = \mathbf{H} / \sum_{u,v}\mathbf{H}(u, v)$ to ensure that all values in $\mathbf{\tilde{H}}$ sums to one.

\subsection{Triangulation}
Given the camera intrinsic $\bm{K} \in \mathbb{R}^{3\times3}$ and extrinsic $\bm{T} \in \mathbb{SE}(3)$ matrices, we can construct the homogeneous projection matrix with the form of
$\bm{M} = \bm{K} \cdot \bm{T}_{[0:3,]} \in \mathbb{R}^{3\times4}$, where $_{[0:3,]}$ indicates the first three rows of $\bm{T}$.
Using this projection matrix $\bm{M}$, the 3D-to-2D projection of the root point $\mathbf{R}$ can be formulated as two independent equations:
\begin{equation}
    \bar{\mathbf{r}} = \bm{M} \cdot \bar{\mathbf{R}}, \rightarrow \begin{cases}
         & u_{\mathrm{r}}(\bm{M}_{[2,]} \bar{\mathbf{R}}) - (\bm{M}_{[0,]}\bar{\mathbf{R}}) = 0, \\
         & v_{\mathrm{r}}(\bm{M}_{[2,]} \bar{\mathbf{R}}) - (\bm{M}_{[1,]}\bar{\mathbf{R}}) = 0,
    \end{cases}
    \label{eq:dlt}
\end{equation}
where $\bar{\mathbf{r}}$, $\bar{\mathbf{R}}$ represents the homogeneous coordinates, \eg $\bar{\mathbf{r}}= (u_{\mathrm{r}}, v_{\mathrm{r}}, 1)^{T}$.
With the $N$ 2D coordinates estimated from $N$ camera views, the \cref{eq:dlt} can be stacked into an overdetermined linear equation $\bm{A} \cdot\bar{\mathbf{R}} = \mathbf{0}$, where $\bm{A} \in \mathbb{R}^{2N\times4}$.
This equation can be solved using the Direct Linear Transform (DLT) \cite[p.312]{hartley2003MVGeometry}. The triangulation result, $\mathbf{R}$, is the singular vector associated with the smallest singular value of $\bm{A}$.

\vspace{-3mm}\section{Point-Embedded Transformer}\label{sec:query_estim}
In the previous stage, we have designed a network branch that can robustly retrieve the absolute root point \emath{\mathbf{R}} from multi-view images.
In this stage, we focus on estimating the relative points \emath{\mathbf{X}_{rel}}, which are offsets relative to the root point.

In the context of multi-view stereo learning method,
the primary focus is to design a network capable of effectively fusing features across different views.
We begin by revisiting the problem setting -- each image captures a frustum space in the camera's field of view, and the common views among all \textit{N} cameras coalesce into a polyhedron, representing the intersection volume of the \textit{N} frustum spaces.
The hand for reconstruction situated within this volume.
As the number of cameras continues to increase, the polyhedron will gradually \textbf{approximates to a sphere} that envelops the hand.
\begin{figure}[!tp]
    \centering
    \includegraphics[width=0.8\linewidth]{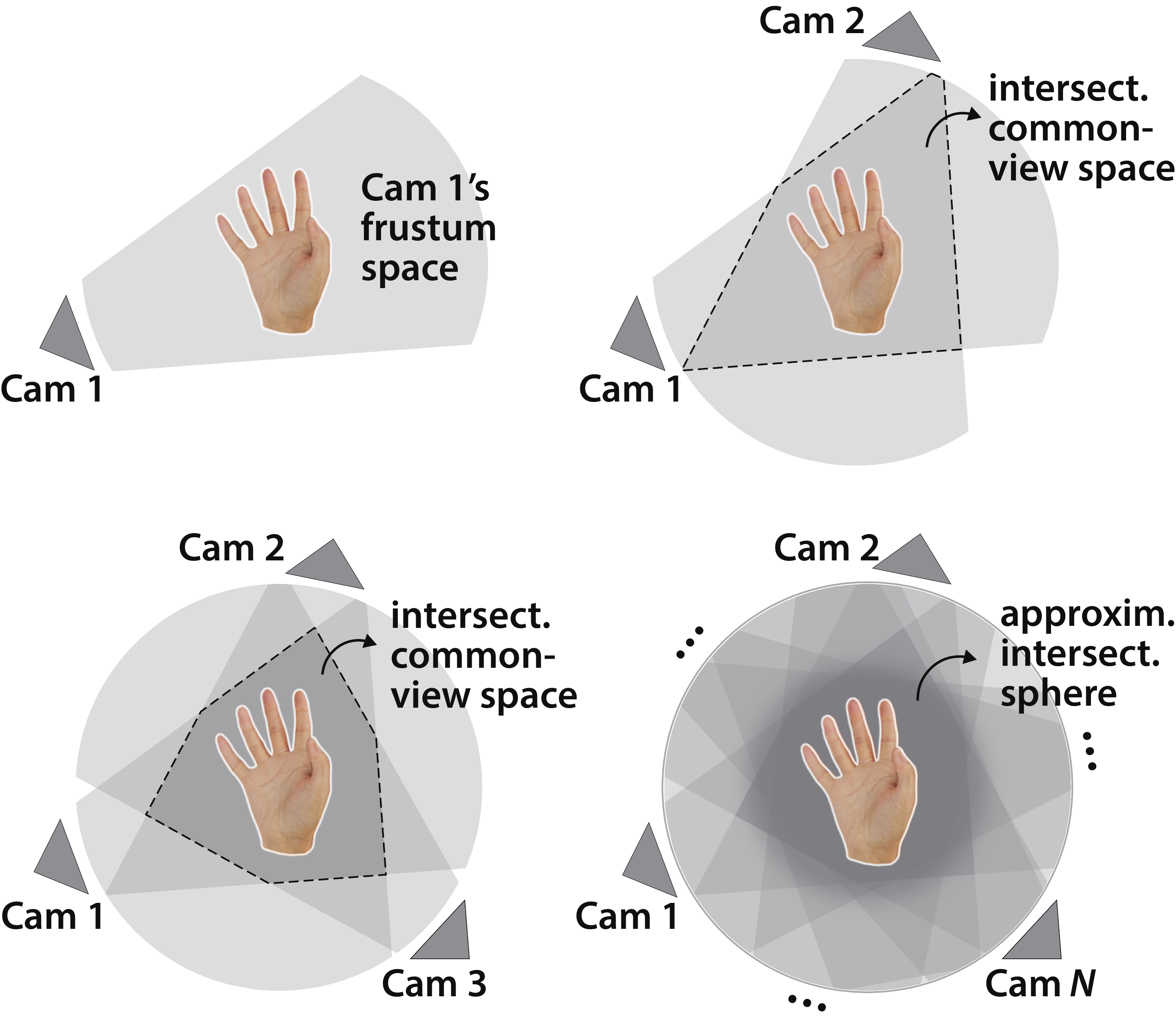}\vspace{-3mm}
    \caption{Illustration of the intersection sphere approximation.}
    \label{fig:sphere_approxim}
\end{figure}

In this context, we propose a novel scheme, wherein the intersection sphere is embedded using a static point cloud. Meanwhile, the hand targeted for reconstruction (\ie the query points \emath{\mathbf{X}}) is treated as a variable point cloud. This enables us to formulate the problem as representing the variable query points through a fixed set of basis points.
To better illustrate this formulation, we take the Fourier Series as an analogy, wherein a complex periodic function can be represented by a series of sinusoidal `basis' functions.
We suggest considering the static point cloud in the intersection sphere as analogous to the basic sinusoidal functions in Fourier Series, and the point cloud of hand as the complex periodic function.

Representing the intersection frustum sphere as point cloud has two advantages:
\begin{enumerate*}[label={\textbf{\arabic*)}}]
    \item the 3D points serve as an ideal medium for carrying spatial relation and image features across multi-views (discussed in \cref{sec:point_embedded}); and
    \item the coordinates of 3D points serves as a natural positional encoding for attention mechanism.
\end{enumerate*}

\vspace{-2mm}\subsection{Point-Embedded Intersection Sphere}\label{sec:point_embedded}
We choose the same strategy as the Basis-Point-Set \cite{prokudin2019bps} to generate the point cloud scattered in the intersection sphere. Specifically, we randomly sampled $M=4096$ 3D points from a ball of diameter $20 cm$\footnote{While the size of intersection sphere varies with multi-view configuration, the sphere of diameter $20cm$ is enough to enclose the human hand. the additional volume beyond this is redundant.} centered at the origin. This point cloud is referred to as the \textbf{Basis Points}, denoted as $\mathbf{P}_{rel} \in \mathbb{R}^{4096\times3}$.
It is important to mention that, although the basis points are randomly sampled, once generated, they remain fixed throughout the training and inference phases.
Given the root point \emath{\mathbf{R}} obtained from the previous stage, we can translate the zero-centerd basis points into the \world space, as $\mathbf{P} = \mathbf{P}_{rel} + \mathbf{R}$.

The next issue is how to map the image features onto the basis points, as they currently exist merely as a set of 3D coordinates. Since the spatial transformation (camera extrinsic, in the form of SE(3) group) from \world space to each camera space is known, we can transform the \emath{\mathbf{P}} into the camera space, and then project these 3D points onto the 2D image plane. At the location of each projection point \emath{\mathbf{p}}, we can gather the image coordinates: $(u, v)$ and the associated image feature $\mathbf{f}$ at this coordinates, as:
\begin{subequations}
    \begin{align}
         & \mathbf{p}_i = (u, v)_i = \bm{K}_i \cdot \bm{T}_i \cdot \mathbf{P}, \label{eq:projection}            \\
         & \mathbf{f}_i = \mathrm{BLI} ( \mathbf{p}_i, ~ \mathrm{CNN}(\mathbf{I}_i) ), \label{eq:feature_align}
    \end{align}
    \label{eq:project_feature}
\end{subequations}
where $i$ denotes the index of $i$-th camera, $\bm{K}_i$, $\bm{T}_i$ is the camera intrinsic and extrinsic matrix, respectively. The term $\mathrm{CNN}(\cdot)$ refers to another image feature extraction network that share the same architecture as the heatmap-based root estimation network in \cref{sec:root_estim}. The $\mathrm{BLI}(\cdot)$ denotes the bilinear interpolation operation that samples the image feature at the location of $\mathbf{p}_i$.

In this way, each basis point is associated with $N$ 2D image coordinates $\{\mathbf{p}_i \}_{i=1}^{N}$ and the corresponding $N$ image features $\{\mathbf{f}_i \}_{i=1}^{N}$.
Based on the established correspondences between these 3D points and their multiple 2D locations, we assert that the spatial relations among cameras in multi-view setup are effectively encoded into these points, endowing these points with a multi-view geometric awareness.
To validate this conclusion, we suppose the transformation of a camera is unknown at this time. Utilizing these 2D-3D correspondences, a Perspective-n-Point (PnP) problem can be constructed, requiring at least three correspondences to solve for the camera's position and orientation (P3P). With 4096 such valid correspondences inherent to the basis points, this is more than adequate to affirm the property of multi-view awareness.

In order to equip the projected feature \emath{\mathbf{f}_{i}} with such awareness, we propose to encode the 2D coordinates \emath{\mathbf{p}_{i}} into the feature \emath{\mathbf{f}_{i}}. This is achieved by leveraging the positional encoding (PE) that is widely used in the vision Transformer.
We use the same 2D ``spatial sine positional encoding'' as in the DETR \cite{Carion2020DETR}.
The final projected feature in $i$-th camera is expressed as:
\begin{equation}
    \mathbf{f}_i = \mathbf{f}_i + \mathrm{PE}(\mathbf{p}_i).
\end{equation}
In the following sections, the term $\mathbf{f}_i$ is used to denote the positional-encoded projected feature.

\vspace{-3mm}\subsection{Projective Aggregation}\label{sec:projective_aggregation}
Each of the projected features $\mathbf{f}_i$ in previous section only contains features from one camera view. To facilitate the fusion of multi-view features, we propose a cross-view feature aggregation mechanism, termed as the Projective Aggregation.
The key operation is to collect the features sampled at the 2D projection of each point in \emath{\mathbf{P}} across $N$ views, and fuse the $N$ independently sampled features into a multi-view feature for that point.

\begin{figure}[!t]
    \centering
    \includegraphics[width=0.8\linewidth]{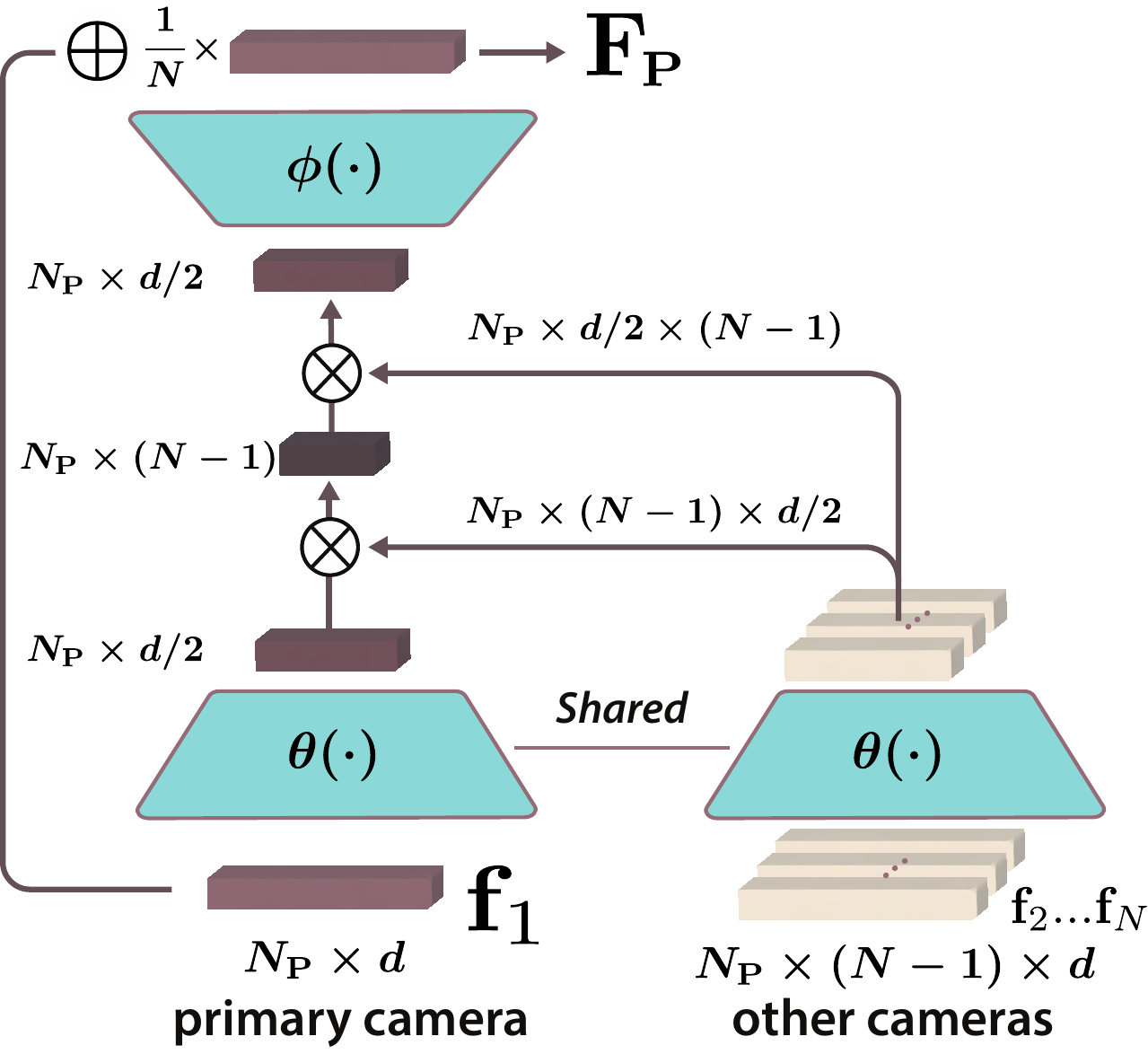}\vspace{-3mm}
    \caption{Architecture of the projective aggregation module. }
    \label{fig:proj_aggregation}
\end{figure}

To achieve the fusion step, we instantiate a non-local neural network featuring a bottleneck architecture.
The network design is illustrated in \cref{fig:proj_aggregation}.
We choose the feature of the first camera (primary camera) as target, with the remaining features from the $N-1$ cameras serving as sources.
The objective is to fuse those source features into the target.
Initially, all $N$ features undergo downsampling to half their original dimension through a shared projection layer.
Then, each source feature independently undergoes a dot-product operation with the target feature, yielding $N-1$ outcomes.
These outcomes are interpreted as weights, and are utilized to perform an element-wise weighted summation on the $N-1$ source features.
The resultant vector is treated as an update for the target feature.
This update is upsampled through an inverse projection layer to restore the original feature dimension.
Thereafter, this update is normalized by a factor of $N$ (number of cameras) -- a step that is necessary to accommodate variations in the number of cameras (number of features to be fused).
At last, the normalized update is added to the target feature to complete the projective aggregation process.
This process is conducted for all basis points, resulting in a set of multi-view \textbf{Basis Features}: $\mathbf{F_P} \in \mathbb{R}^{4096\times d}$, where $d$ is the feature dimension.
When only one camera is available, the aggregation mechanism is disabled. In such case, the projected features $\mathbf{f}_i$ is directly passed to the next stage.
Adding this single-view branch during training force that feature from each view contains sufficient information to reconstruct the hand.
The projective aggregation can be summarized as:
\begin{equation}
    \mathbf{F_P} = \begin{cases}
        \mathbf{f}_1 + \mathcal{G}(\mathbf{f}_1, ..., \mathbf{f}_N), & \text{if } N > 1, \\
        \mathbf{f}_1,                                                & \text{if } N = 1,
    \end{cases}
    \label{eq:proj_aggr}
\end{equation}
where $\mathcal{G}(\cdot)$ denotes the bottleneck non-local network, as:
\begin{equation}
    \mathcal{G}(\mathbf{f}_1, ..., \mathbf{f}_N)  =  \frac{1}{N} \phi \left( \sum^{N}_{j=2} \theta(\mathbf{f}_1)^{T}\theta(\mathbf{f}_j) \cdot \theta(\mathbf{f}_j)\right),
\end{equation}
and the $\theta$, $\phi$ are the learnable projection and inverse projection layers, respectively.

\subsection{Learnable Query Points}\label{sec:learnable_query}
\model follows the standard Transformer architecture to predict the root-relative position of the query points \emath{\mathbf{X}_{rel} \in \mathbb{R}^{799\times3}}.
The Transformer decoder transforms 799 embeddings, each of dimension $d$,  using the multi-head attention mechanism. Given that attention is permutation-invariant, it is important that each of the 799 query point embeddings be distinct.
To this end, we randomly initialize a set of learnable positional encoding, termed as \textbf{Query Embeddings}: \emath{\mathbf{E_X} \in \mathbb{R}^{799\times d}}.
These query embeddings are fed to the Transformer decoder to serve as the input for the first decoder layer.

In addition to the query embeddings, the 3D coordinates of the query points are also subject to learning.
We initialize the query points using consistent joint and vertex position derived from a zero-pose and mean-shape MANO hand template, denoted as \emath{\mathbf{X}^{\dagger}_{rel}}.

It is worth to mention that, although both the query points, \emath{\mathbf{X}_{rel}}, and query embeddings, \emath{\mathbf{E_X}}, are learnable, they are treated differently within the model.
After each iteration of the model training, the query embedding will be updated via back-propagation; however, the initial query points remain fixed at the template positions.
After trained to convergence, the model preserves the learned query embeddings as constant during inference, whereas the query points continue to be initialized at the template positions. For each multi-view sample, the model updates the initial query points to the final prediction.

\subsection{Point-Embedded Transformer Decoder}\label{sec:point_transformer}
\begin{figure}[!t]
    \centering
    \includegraphics[width=0.9\linewidth]{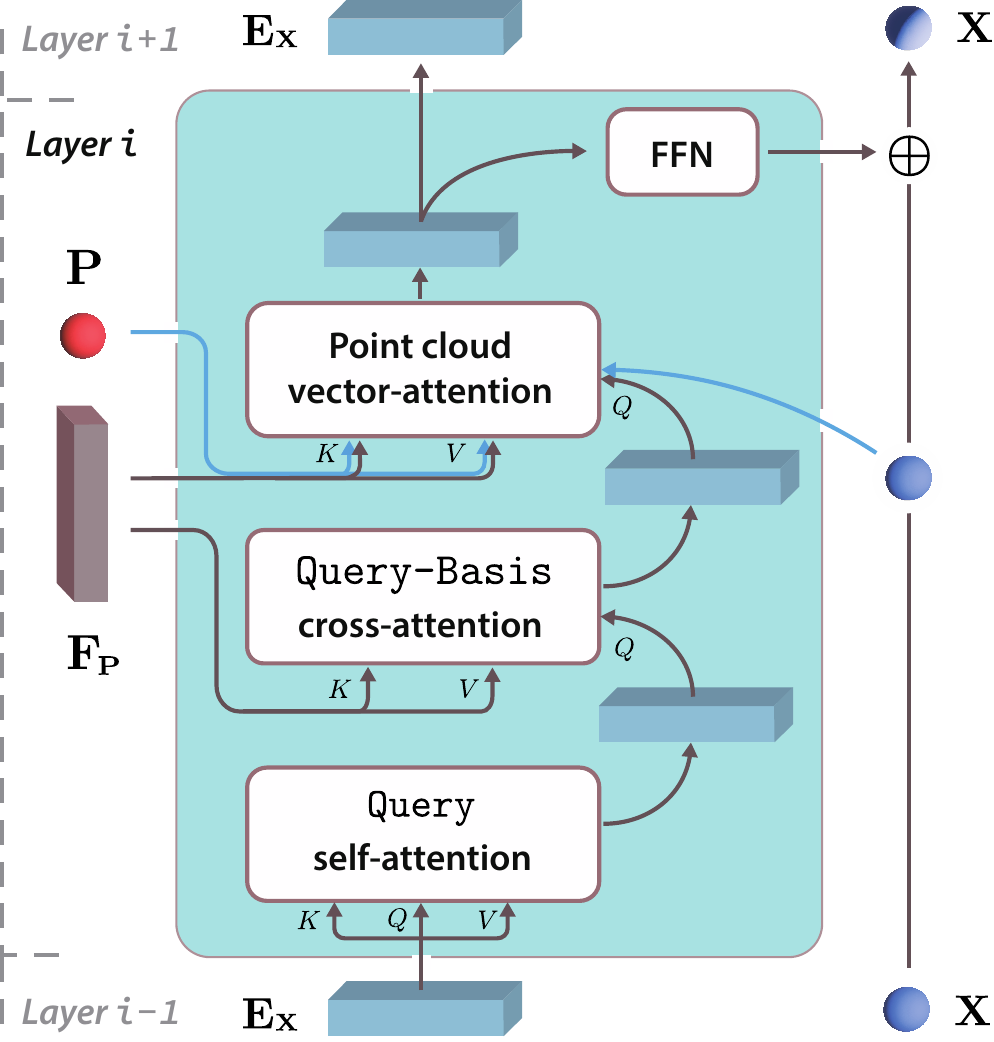}
    \caption{One layer of the Point-Embedded Transformer Decoder.}
    \label{fig:poemb_transformer}
\end{figure}

Given two point clouds in the world space:
\begin{itemize}
    \setlength\itemsep{2pt}
    \item \textbf{\Basis}: $ (\mathbf{P},~~ \mathbf{F_P} )$, \\ representing the basis points $\mathbf{P} \in \mathbb{R}^{4096\times3}$ and basis features $\mathbf{F_P}\in \mathbb{R}^{4096\times d}$; and
    \item \textbf{\Query}: $ (\mathbf{X}, ~\mathbf{E_X} )$, \\representing the query points $\mathbf{X} \in \mathbb{R}^{799\times3}$ and query embeddings $\mathbf{E_X}\in \mathbb{R}^{799\times d}$,
\end{itemize}
\model employs a decoder-only Transformer architecture to progressively transform multi-view features from basis points to query points, and to update the 3D position of the query points.
\cref{fig:poemb_transformer} illustrates the architecture of the point-embedded Transformer.
It consists of a stack of $L$ decoder layers, where each layers contains four modules, including a multi-head self-attention, a multi-head cross-attention, a point cloud vector attention, and a feed-forward network.

\paraheading{\Query's Self-Attention.}
Given the query embedding as an input sequence  of length 799: $\mathbf{E_X} = [\mathbf{E_{X}}_1, \mathbf{E_{X}}_2, ..., \mathbf{E_{X}}_{799}]$, we first conduct a multi-head self-attention on these embedding sequence.
Each self-attention function first projects this sequence into three different components: \ie $Q$, $K$, and $V \in \mathbb{R}^{799\times d}$, as:
\begin{equation}
    Q, K, V = f_Q(\mathbf{E_X}), f_K(\mathbf{E_X}), f_V(\mathbf{E_X}),
    \label{eq:qkv}
\end{equation}
where the $f_Q$, $f_K$, $f_V$ are three linear projection layers.
The output of a self-attention is computed as:
\begin{equation}
    \mathrm{Atten}(\mathbf{E_X}) = \mathrm{softmax}\left(\frac{QK^T}{\sqrt{d}}\right)V.
\end{equation}
Since all $Q, K, V$ originate from an identical input embeddings, this \Query's self-attention captures the internal relationships among the embeddings of different query points.
The $\mathrm{Atten}(\mathbf{E_X})$ is used to update the query embeddings and passed to the next module.

\paraheading{\Query-\Basis Cross-Attention.}
Given the basis features $\mathbf{F_P}$ as another input sequence of length 4096, we conduct a multi-head cross-attention between the query embeddings and the basis features.
The cross-attention is achieved by substituting the input of $f_K(\cdot)$, $f_V(\cdot)$ in \cref{eq:qkv} with the basis features $\mathbf{F_P}$.
This cross-attention mechanism enables the query embedding to interact with the basis features at the full-set level.

\paraheading{Point Cloud Vector Attention.}
In order to leverage the power of points, we inject a point cloud attention to enhance the local interaction between \Query and \Basis.
Point cloud is invariant to the permutation of points and primarily subjected to the relative distance among points.
Thus, for each point in \Query, our mechanism adeptly merges information from neighboring points within \Basis, based on their proximal relations.
Our point cloud attention is build upon the principles established by the Point Transformer \cite{Zhao2021PointT}. Unlike the standard scalar attention, which calculates the attention from the dot-product matrix ($QK^T$ in \cref{eq:qkv}), the point cloud attention is a vector attention, where the attention is computed from the \textbf{subtraction vector}. Supposing $\mathbf{S}_i$ is a subset of basis points \emath{\mathbf{P}} that encompasses the $k$ nearest neighbor of the $i$-th query point \emath{\mathbf{X}_i}. The vector attention is computed as:
\begin{equation}
    \footnotesize
    \mathbf{E_X}_i = \sum_{\forall \mathbf{P}_j  \in \mathbf{S}_i} \mathrm{softmax}\Big(\gamma\big(\underbrace{\alpha(\mathbf{E_X}_i) - \beta(\mathbf{F_P}_j)}_{\text{subtraction vector}} + \delta \big)\Big) \odot (\psi(\mathbf{F_P}_j) + \delta).
    \label{eq:vector_atten}
\end{equation}
The $\delta$ is the positional encoding of a query point $\mathbf{X}_i$ with respect to a basis point $\mathbf{P}_j$, as $\delta = \xi(\mathbf{X}_i - \mathbf{P}_j)$.
The $\alpha, \beta, \gamma, \psi, \xi$ are five learnable embedding networks.

\paraheading{Feed-Forward Network.} After the query embeddings $\mathbf{E_X}$ undergo sequential processing through the self-attention, cross-attention, and vector attention module, the updated embeddings are input to a feed-forward network (FFN) to regress the update on the coordinates of the root-relative query points $\mathbf{X}_{rel}$, as
\begin{equation}
    \mathbf{X}_{rel} = \mathbf{X}_{rel} + \mathrm{FFN}(\mathbf{E_X}).
\end{equation}
The root point $\mathbf{R}$ is added to the output query points $\mathbf{X}_{rel}$ to obtain the final 3D coordinates of the query points $\mathbf{X}$.
Subsequently, the updated query points $\mathbf{X}$ and query embeddings $\mathbf{E_X}$ are propagated to the next layer to serve as inputs.

\subsection{Misc. Modifications on the POEM Model}
The POEM model in this journal version (referred to as \textbf{\POEMg}) includes two additional technical modifications over the conference version (\textbf{\POEMs}) \cite{yang2023poem}:

\paraheading{i) Output of the First Stage}: In version 1 -- \textbf{\POEMs}, all 21 joints $\mathbf{J}$ are triangulated from heatmaps produced in the first stage, and these joint positions initialize the vertices. This design relegates part of the query points estimation to the first, less powerful stage, resulting in potential inaccuracies in extreme cases (\eg, with fewer cameras) where joints determined via DLT can be imprecise. In version 2 -- \textbf{\POEMg}, we merge vertices and joints into the query points $\mathbf{X}$, deferring their prediction to the second stage (Point-Embedded Transformer). The first stage now only outputs the root position, specifically the MCP joint (joint ID = 9). This adjustment does not alter the two-stage modeling approach; the second stage continues to place the Basis Points Set (BPS) according to the root estimation.

\paraheading{ii)Basis Points Acquisition}: In \textbf{\POEMs}, the 4096 BPS points are sampled via Ball Query \cite{qi2017pointnet++} from the mesh grid points of multi-cameras, with the root point as the origin. The coordinates of the mesh grid points are generated using a linear-increasing discretization (LID) strategy for the Z-axis, resulting in varying sampling densities -- denser near the camera and sparser at greater depths. In \textbf{\POEMg}, we generate 4096 points via Gaussian sampling within a sphere and then translate the origin to the root point, preserving their relative positions. This approach avoids the need for Ball Query sampling during each forward pass, and promotes consistent point density across all views.

These modifications do not change the core methodology proposed in this paper. Therefore, their impact is not separately analyzed.

\section{Experiments}\label{sec:experiments}

The experiment section of this paper centers on the evaluation under four configurations:

\paraheading{i) Model Architecture Analysis:} We assess the performance and advantages of the various components within the \model's architecture.
This setup corresponds to the experiments presented in the conference version of this work (in CVPR 2023) \cite{yang2023poem}. The specific model trained under this configuration is designated as \textbf{\POEMs}.

\paraheading{ii) Scalability Assessment:} a POEM model is designed to scale in terms of both the parameters count and data used for training, and to adapt in scenario where number of cameras and their poses are varied. This model is trained on five multi-view datasets plus one monocular dataset in a randomly mixing patterns and evaluated on each single test set to assess its performance. The model trained in this configuration is referred to as \textbf{\POEMg} (corresponds to the version in this manuscript).

\paraheading{iii) Feature Dimension Analysis:} a \POEMg model is further adapted to encompass three different feature dimension -- small, base, and large -- to examine the correlation between model size and performance.

\paraheading{iv) Real-world Application:} The trained \POEMg model is put to the test on a self-collected real-world dataset to further evaluate the model's generalizability on out-of-domain data.

\subsection{Implementation Details}\label{sec:implem}
All the experiments related to the \textbf{\POEMs} and its comparative alternatives use the ResNet34 \cite{resnet} as the CNN back bone. Practically, same ResNet convolution layers are shared between the CNN for heatmap estimation and feature extraction, with different deconvolution layers assigned to each task. Within \POEMs Transformer decoder, the Query-Basis cross-attention is omitted. The decoder layers $L$ are set to 6, the hidden feature dimension $d$ is 256, and the number $k$ of nearest neighbors is 16.
The Transformer decoder is initialized with xavier uniform distribution and the CNN backbone is initialized with ImageNet \cite{imagenet} pre-trained weights.
This \POEMs model is independently trained on each multi-view dataset with a predetermined number and order of cameras, thereby guaranteeing a consistent configuration of camera poses throughout the training and testing phases.
The model is trained using the Adam optimizer with a learning rate of 1e-4. The learning rate is decayed by a factor of 0.1 at the 70\% of the total training epochs. The batch size is set to 16, and the model is trained for 100 epochs.

The generalizable \textbf{\POEMg} model leverage the HRNet \cite{wang2020deep} as the CNN backbone. Same as the \POEMs, the HRNet's convolution layers are shared by the CNN for heatmap estimation and feature extraction.
The \POEMg Transformer decoder adopts the Query-Basis cross-attention mechanism to increase the holistic interaction between the two point clouds.
The decoder layers $L$ are set to 3 and the number $k$ of nearest neighbors is 32.
The hidden feature dimension $d$ takes four different values: 128, 256 and 512, corresponding to the three variants of the \POEMg model: \POEMg-small, \POEMg-base (default), and \POEMg-large, respectively. \POEMg model is trained on a mixed dataset that combines five multi-view datasets and one monocular hand dataset. The training data is randomly mixed with varying numbers of cameras (from 1 to 8) and varying orders of different cameras.
The model is trained using the Adam optimizer with an initial learning rate of 1e-4. The learning rate is decayed via a cosine annealing scheduler after each back-propagation step. The batch size is set to 32, and the model is trained for 10 epochs (total 2,100,000 iterations) on the mixed dataset.
The training of \POEMg models are conducted on 4 NVIDIA A6000 GPUs. The training duration of \POEMg-small, \POEMg-base, and \POEMg-large model span 102, 116, and 152 hours, respectively.

\subsection{Datasets}\label{sec:datasets}
To train the \POEMg model, we integrate multiple datasets that provide the observation and 3D annotation of hand from multiple calibrated cameras.
These dataset include DexYCB \cite{Chao2021DexYCB}, HO3D \cite{hampali2020ho3dv2}, OakInk \cite{Yang2022OakInk}, ARCTIC \cite{fan2023arctic}, and InterHand2.6M \cite{moon2020interhand2}.
Additionally, we incorporate one monocular hand dataset, FreiHAND \cite{zimmermann2019freihand}, to accommodate the single-view scenario in our training regimen (refer to the case where $N=1$ in \cref{eq:proj_aggr}).

For the purpose of establishing a standardized multi-view framework, for any given timestamp, we compile a set of multi-view images from all cameras. This assembly of images, along with their respective annotations, constitutes a single multi-view frame. For all the datasets, we filter out samples featuring left hand. We designate this version of the dataset with the specific multi-view configuration as `\textbf{-Mv}'.

\paraheading{DexYCB-Mv.}
We adhere to the `S0' split delineated by the original dataset for training, validation, and testing subsets. Each multi-view frame in the DexYCB-Mv dataset comprises \textbf{8} images. The training set contains 25,387 multi-view frames, totaling 203,096 images when considering individual cameras. The validation set contains 1,412 multi-view frames, while the test set includes 4,951 multi-view frames.

\paraheading{HO3D-Mv.}
We select seven sequences from the HO3D dataset that provide complete observations from all \textbf{5} cameras. For our training set, we chose sequences named as `ABF1', `BB1', `GSF1', `MDF1', and `SiBF1', whilst designating the remaining `GPMF1' and `SB1' sequences for the testing set. In total, the \hodddmv training set is comprised of 9,087 multi-view frames, translating to 45,435 images when considering individual camera. The testing set contains a sum of 2,706 multi-view frames.

\paraheading{OakInk-Mv.}
The OakInk-Mv dataset is curated in alignment with the official `SP2' object-based split. Each multi-view frame within OakInk-Mv compiles images from all \textbf{4} cameras. Approximately one-fourth of the sequences within OakInk exhibit scenarios of two individuals engaging in an object handover, thereby capturing both hands within a single image. For these particular sequences, we opt for independent training and evaluation of each hand. In its entirety, the OakInk-Mv comprises 58,692 multi-view frames in the training set and 19,909 multi-view frames in the testing set.

\paraheading{ARCTIC-Mv.}
The ARCTIC-Mv dataset is constructed by compiling images from all \textbf{8} cameras in the official 'P1' split. The training set consists of 135,271 multi-view frames, equivalent to 1,082,168 images when considering individual camera. The testing set contains 17,392 multi-view frames.

\paraheading{InterHand-Mv.}
Most samples in the original InterHand-2.6M dataset consists of more than 50 cameras. To create a smaller multi-view variant of this dataset, we employ an iterative process of randomly selecting \textbf{8} views from the whole observations. This approach ensures optimal utilization of the available data. The Interhand-Mv dataset encompasses 210,006 multi-view samples for training, equivalent to 1,680,048 monocular samples, and 85,255 multi-view samples for the testing set.

\paraheading{FreiHAND.}
Within our framework, we treat the training sample from FreiHAND as a particular instance of a multi-view sample where the number of views equals one. The FreiHAND dataset comprises a total of 32,560 training frames.

\paraheading{The Overall Training Data of \POEMg.}
The overall training data of \POEMg model is the mixture of the above six datasets.
The mixture ratio of the datasets — DexYCB-Mv, HO3D-Mv, OakInk-Mv, ARCTIC-Mv, InterHand-Mv, and FreiHAND — is set to 0.18, 0.18, 0.18, 0.18, 0.18, and 0.1, respectively.
For each training sample, we randomly select from one to the max numbers of observations and randomly shuffle the order of the selected observations.
In the resulting multi-view configuration, we anchor the world coordinate system to the camera in the first position, and calculate the transformation (extrinsic matrix) of the subsequent cameras with respect to this primary camera.

\subsection{Metrics}\label{sec:metric}
To evaluate the performance of our proposed model, we report the Mean Per Joint Position Error (MPJPE) and the Mean Per Vertex Position Error (MPVPE), both measured in the camera space with unit millimeters ($mm$).
Additionally, we analyze both the MPJPE and MPVPE within two reference system: a root-relative (RR) system and a system after conducting Procrustes Analysis (PA).
The `RR' suggests ignoring the drift on root translation. The `PA' provides an assessment that disregards variations in scale, rotation, and translation, thus offering a measure of shape alignment independent of pose.
Additionally, we calculate the Area Under the Curve (AUC) of the percentage of correct keypoint predictions across various landmark thresholds. The AUC effectively encapsulates the model's discriminative capacity on localizing keypoint across a specified spectrum of precision thresholds.

\subsection{Data Augmentation}\label{sec:implem}
To train the \model model, we apply standard image augmentation techniques including random center offset, scaling, and color jittering.
In the context of image-to-3D reconstruction tasks, integrating random rotation augmentations is essential for enhancing the model's resilience to orientation changes.
Different from the rotation in a single-view scenario - where a rotation on the image corresponds to the same rotation on the 3D hand model - the rotation augmentation in multi-view must not change the orientation of 3D hand model in the world space.
To explain this augmentation in multi-view setting, when an input image is augmented by a rotation of an angle $a$, this transformation equates to left-multiplying the existing rotation to the camera's extrinsic matrix within the projection formula \cref{eq:projection}
\begin{equation}
    \mathrm{rot}(a)\cdot \mathbf{p} =  \bm{K}\cdot {\mathrm{Rot}(a)\cdot\bm{T}} \cdot \mathbf{P},
    \label{eq:rotate}
\end{equation}
where $\mathrm{rot}(a) \in \mathbb{SO}(2)$ and  $\mathrm{Rot}(a) \in \mathbb{SO}(3)$ denotes the rotation matrix in 2D and 3D space, respectively. In this formula, the $\big( {\mathrm{Rot}(a)\cdot\bm{T}} \big)$\footnote{The matrix $\mathrm{Rot}(a)$ is a 4x4 transformation matrix belonging to the $\mathbb{SE}(3)$ group, specifically exhibiting a zero translation component.} represents the new camera extrinsic after augmentation.
In this way, the 3D hand model remains invariant in the world space, thereby allowing us to apply random augmentations to images from different views independently.

\subsection{Main Experiments}\label{sec:main_exp}
\begin{table*}[t]
    \addtolength{\tabcolsep}{-1pt}
    \begin{center}
        \caption{Quantitative results in the main experiments.}\vspace{-3mm}
        \label{table:quantitative}
        \footnotesize
        \begin{tabular}{lllcccccccc}
            \toprule
            datasets                   & \#          & methods                                                 & \MPVPE         & \RRV           & \PAV          & \AUCVtwenty   & \MPJPE         & \RRJ           & \PAJ          & \AUCJtwenty   \\
            \midrule
            \multirow{5}*{{DexYCB-Mv}} & \texttt{1}  & \textbf{\POEMs}                                         & \textbf{6.13}  & \textbf{7.21}  & \textbf{4.00} & \textbf{0.70} & \textbf{6.06}  & \textbf{7.30}  & \textbf{3.93} & \textbf{0.68} \\
                                       & \texttt{2}  & MvP\cite{Wang2021MVP}                                   & 9.77           & 12.18          & 8.14          & 0.53          & 6.23           & 9.47           & 4.26          & 0.69          \\
                                       & \texttt{3}  & PE-MeshTR \cite{Liu2022PETR,lin2021metro}               & 7.41           & 8.67           & 4.70          & 0.64          & 7.49           & 8.87           & 4.76          & 0.64          \\
                                       & \texttt{4}  & FTL-MeshTR \cite{Remelli2020LightweightM3,lin2021metro} & 8.75           & 9.80           & 5.75          & 0.59          & 8.66           & 9.81           & 5.51          & 0.59          \\
                                       & \texttt{5}  & iter. optim. using \cite{chen2021cmr}                   & 7.33           & 8.71           & 5.29          & 0.65          & 7.22           & 8.77           & 5.19          & 0.65          \\
            \midrule
            \multirow{4}*{OakInk-Mv}   & \texttt{6}  & \textbf{\POEMs}                                         & \textbf{6.20}  & \textbf{7.63}  & \textbf{4.21} & \textbf{0.70} & \textbf{6.01}  & \textbf{7.46}  & \textbf{4.00} & \textbf{0.69} \\
                                       & \texttt{7}  & MvP \cite{Wang2021MVP}                                  & 9.69           & 11.75          & 7.74          & 0.53          & 7.32           & 9.99           & 4.97          & 0.64          \\
                                       & \texttt{8}  & PE-MeshTR \cite{Liu2022PETR,lin2021metro}               & 8.34           & 9.67           & 5.75          & 0.60          & 8.18           & 9.59           & 5.42          & 0.61          \\
                                       & \texttt{9}  & FTL-MeshTR \cite{Remelli2020LightweightM3,lin2021metro} & 9.28           & 10.88          & 6.61          & 0.56          & 8.89           & 10.66          & 6.01          & 0.58          \\
            \midrule
                                       &             &                                                         &                &                &               & \AUCVfifty    &                &                &               & \AUCJfifty    \\
            \multirow{4}*{HO3D-Mv}     & \texttt{10} & \textbf{\POEMs}                                         & \textbf{17.20} & \textbf{21.45} & \textbf{9.97} & \textbf{0.66} & \textbf{17.28} & \textbf{21.94} & \textbf{9.60} & \textbf{0.63} \\
                                       & \texttt{11} & MvP \cite{Wang2021MVP}                                  & 20.95          & 27.08          & 10.04         & 0.59          & 18.72          & 24.90          & 10.44         & 0.60          \\
                                       & \texttt{12} & PE-MeshTR \cite{Liu2022PETR,lin2021metro}               & 23.49          & 29.19          & 11.31         & 0.55          & 23.94          & 30.23          & 11.67         & 0.54          \\
                                       & \texttt{13} & FTL-MeshTR \cite{Remelli2020LightweightM3,lin2021metro} & 24.15          & 33.53          & 10.56         & 0.53          & 24.66          & 34.74          & 10.76         & 0.52          \\
            \bottomrule
        \end{tabular}
    \end{center}
\end{table*}

\model targets on single hand reconstruction under the multi-view RGB observations. We find that the Multi-view Pose Transformer (MvP) \cite{Wang2021MVP}, which regress the mesh parameters catered to a skinning body model (\eg MANO\cite{romero2017embodied}) using vision Transformer, is an ideal baseline for our model. MvP also employs another module to associate keypoints among different persons. In our experiments, we only compare \model with MvP in the scenario of single-hand reconstruction.

Except for the MvP model, we also simulate several POEM's alternative designs for comparison.
The purpose of the simulation models is to combine the state-of-the-art architecture in monocular hand mesh reconstruction with the advanced fusion scheme adopted in common multi-view settings.

In terms of monocular hand mesh reconstruction, the Mesh Transformer: METRO \cite{lin2021metro} is a representative method that effectively adopts the Transformer decoder to directly regress all vertex position on the hand mesh. In METRO, vertices for estimation are treated as the token of the Transformer decoder. These vertices token are initialized by the features from the backbone image encoder concatenates with the vertex coordinates at its template position, and updated to form the final position of the hand mesh.
We adopt similar design in the simulation models.
In METRO, the Transformer decoders only employ the self-attention, which means the vertices token only interact with themselves. In the simulation models, we adopts a more general decoder design -- the decoder from DETR, where the image features from the backbone encoder are treated as the Key and Value token for the attention mechanism, and Query token is the query embeddings $\mathbf{E_X}$ (as described in \cref{sec:learnable_query}). At the output stage of the Transformer decoder, the query embeddings with dimension of $799 \times d$ are converted to the vertices position with dimension of $799 \times 3$ by a MLP layer.

In terms of the multi-view fusion scheme, we investigate two types of fusion strategies:
\begin{enumerate*}[label=\textbf{\roman*)}.,leftmargin=15pt]
    \item Position Embedded fusion and
    \item Canonical Fusion.
\end{enumerate*}

\paraheading{i) Position Embedded Fusion.} Position Embedding Transformer (PETR) \cite{Liu2022PETR} proposes using the Position Embedding to integrate feature from multi-views. It discretizes the camera frustum space using a mesh grid and embeds the world position of each grid into the image feature.
We adopt this design in one alternative model, \textbf{PE-MeshTR}. Specifically, a mesh grid $\bm{M}_{cam}$ is a tensor of shape $(W, H, D, 3)$, wherein each grid stores a 3D point coordinates along a camera ray vector. Leveraging the camera extrinsic matrix $\bm{T}$, we can transform the mesh grid from the camera space to the world space: $\bm{M}_{\mathcal{W}} = \bm{T}^{-1} \cdot \bm{M}_{cam}$. Meantime, the image feature $\mathbf{B}_{cam}$ is a tensor with shape: $(W, H, C)$, where $C$ is the numbers of the feature's channel. The fusion process consists of an alignment stage and a embed stage.
In the alignment stage, the image feature is reshaped to $\mathbf{B}_{cam}^{\wedge}$: $(W, H, D, \frac{C}{D})$ to aligns its dimensions with the mesh grid, where operation $\wedge$ and $\vee$
denote the up-reshaping and down-flattening operations, respectively.
In the embed stage, the position encoding of $\bm{M}_{\mathcal{W}}$: $f_M(\bm{M}_{\mathcal{W}})$ is embedded into the image feature via:
\begin{equation}
    \mathbf{B}_{\mathcal{W}} = (~ \mathbf{B}_{cam}^{\wedge} + f_M(\bm{M}_{\mathcal{W}}) ~)^{\vee},
\end{equation}
where $f_M$ is a series of sinusoids function.
This process yields $\mathbf{B}_{\mathcal{W}}$, which we refer to as the \textbf{position-embedded image feature}.
The  $\mathbf{B}_{\mathcal{W}}$  is then flattened into one dimension with shape $(W \times H \times C)$.
After flatten, $N$ position embedded features from $N$ cameras are concatenated into a single tensor, and fed into the Transformer decoder.

\paraheading{ii) Canonical Fusion.} Feature Transform Layer, initially proposed in \cite{worrall2017interpretable}, applies inverse spatial transformations directly to image features to disentangle the learned feature representation from the transformations. Subsequent work on multi-view pose estimation \cite{Remelli2020LightweightM3, Han2022UmeTrack} adopt this design, demonstrating its effectiveness in 3D space. When the inverse of the camera pose is applied to image features, these features are referred to as canonical. Consequently, fusing canonical features is called Canonical Fusion.
We adopt this design in another alternative model, \textbf{FTL-MeshTR}. Specifically, given the image feature $\mathbf{B}_{cam}$, we first reshape it to $\mathbf{B}_{cam}^{\wedge}$: $(W, H, \frac{C}{3}, 3)$. Since the last dimension has the shape of $(3)$, this $\mathbf{B}_{cam}^{\wedge}$ can be treated as a 3D point-set in its camera space.
The canonical fusion is conducted in two stages: canonicalization and fusion.
In the first stage, the inverse camera extrinsic matrix is applied to the point-set to transform it to the world space:
\begin{equation}
    \mathbf{B}_{\mathcal{W}} = (~\bm{T}^{-1} \cdot \mathbf{B}^{\wedge}_{cam}~)^{\vee},
\end{equation}
This $\mathbf{B}_{\mathcal{W}}$ is referred to as the \textbf{canonical image feature}, indicating the camera pose is disentangled from the image feature.
Then, in the fusion stage, the resulting $\mathbf{B}_{\mathcal{W}}$ from $N$ cameras are concatenated into a single tensor, and fed into a convolution layer to fuse the canonical features across different views (different channels). Finally, this fused canonical feature is flattened into one dimension with shape $(N\times W \times H \times C)$ and input into the Transformer decoder.

\bigskip
\begin{figure}[!t]
    \centering
    \includegraphics[width=1.0\linewidth]{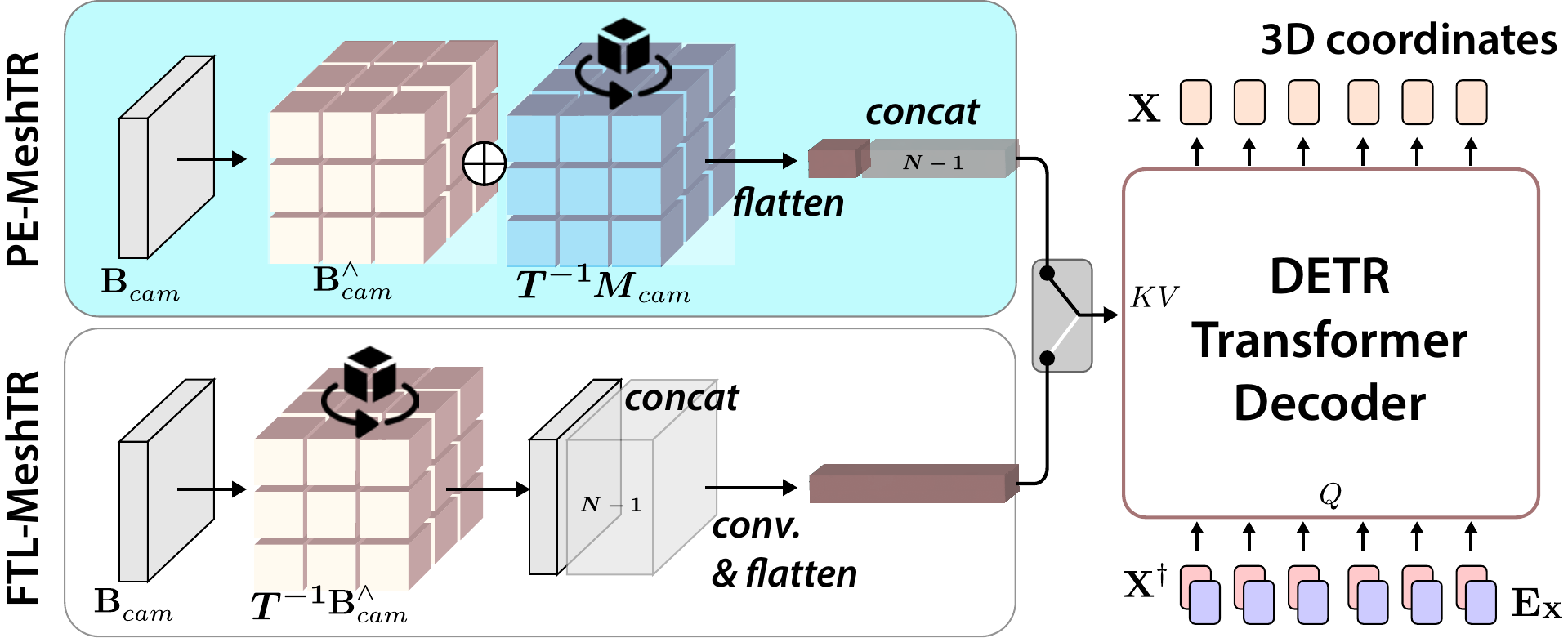}\vspace{-2mm}
    \caption{Architecture of the PE-MeshTR and FTL-MeshTR.}
    \label{fig:sim_models}
\end{figure}
The architecture of these two models are illustrated in \cref{fig:sim_models}.
We train the \POEMs and these alternative models on the \dexycbmv, \hodddmv, and \oakinkmv datasets individually, and evaluate their performances on the corresponding test sets.
The evaluation metrics are reported in \cref{table:quantitative}. Our \model model outperforms these methods across all metrics.

\begin{table*}[!htbp]
    \addtolength{\tabcolsep}{-1pt}
    \begin{center}
        \caption{Quantitative results of the the ablation study.}\vspace{-3mm}
        \label{tab:ablation}
        \footnotesize
        \begin{tabular}{lllcccccccc}
            \toprule
            datasets              & \#          & methods                                         & \MPVPE         & \RRV           & \PAV          & \AUCVtwenty   & \MPJPE         & \RRJ           & \PAJ          & \AUCJtwenty   \\
            \midrule
            \multirow{5}*{DexYCB-Mv} & \texttt{1}  & \textbf{\POEMs}                                 & \textbf{6.13}  & \textbf{7.21}  & \textbf{4.00} & \textbf{0.70} & \textbf{6.06}  & \textbf{7.30}  & \textbf{3.93} & \textbf{0.68} \\
                                  & \texttt{2}  & \POEMs-1stage                                   & 9.42           & 10.06          & 5.84          & 0.61          & 9.27           & 9.69           & 5.22          & 0.60          \\
                                  & \texttt{3}  & \POEMs-w/o-proj                                 & 6.57           & 7.69           & 4.42          & 0.68          & 6.54           & 7.82           & 4.37          & 0.67          \\
                                  & \texttt{4}  & \POEMs-w/o-pt                                   & 7.63           & 8.94           & 5.48          & 0.63          & 7.20           & 8.58           & 4.89          & 0.65          \\
            \midrule
            \multirow{5}*{OakInk-Mv} & \texttt{6}  & \parbox{1.3cm}{\textbf{\POEMs}}(\textbf{V-emb}) & \textbf{6.20}  & \textbf{7.63}  & \textbf{4.21} & \textbf{0.70} & \textbf{6.01}  & \textbf{7.46}  & \textbf{4.00} & \textbf{0.69} \\
                                  & \texttt{7}  & \parbox{1.3cm}{~}(G-emb)                        & 6.25           & 7.65           & 4.33          & 0.70          & 6.05           & 7.66           & 4.10          & 0.69          \\
                                  & \texttt{8}  & \parbox{1.3cm}{~}(G\&V-emb)                     & 6.23           & 7.63           & 4.30          & 0.70          & 6.05           & 7.65           & 4.09          & 0.69          \\
            \noalign{\smallskip}
                                  & \texttt{9} & \POEMs-w/o-proj                                 & 6.42           & 7.82           & 4.50          & 0.69          & 6.25           & 7.84           & 4.28          & 0.68          \\
            \midrule
                                  &             &                                                 &                &                &               & \AUCVfifty    &                &                &               & \AUCJfifty    \\
            \noalign{\smallskip}
            \multirow{4}*{HO3D-Mv}   & \texttt{10} & \textbf{\POEMs}                                 & \textbf{17.20} & \textbf{21.45} & \textbf{9.97} & \textbf{0.66} & \textbf{17.28} & \textbf{21.94} & \textbf{9.60} & \textbf{0.63} \\
                                  & \texttt{11} & \POEMs-w/o-proj                                 & 18.83          & 22.26          & 10.83         & 0.63          & 18.48          & 22.73          & 10.39         & 0.63          \\
                                  & \texttt{12} & \POEMs-w/o pt                                   & 19.26          & 24.32          & 12.45         & 0.62          & 18.20          & 23.80          & 10.56         & 0.63          \\
            \bottomrule
        \end{tabular}
    \end{center}
\end{table*}

In addition to comparing the \model with these end-to-end multi-view reconstruction methods, we also compare it with an iterative optimization method.
Fitting a 3D hand model to align with its 2D keypoint estimations from different views is a common practice for obtaining 3D ground-truth in hand datasets.
For instance, the HOnnotate method \cite{hampali2020ho3dv2} aggregates visual cues such as 2D hand keypoints and 2D segmentation from pre-trained networks and implements an iterative optimization to curate the HO3D dataset.
In a similar way, we construct our fitting objectives in three parts:
\begin{enumerate}[label=\roman*).,leftmargin=15pt]
    \item The projection of the 3D hand joints should match the 2D keypoint estimations in each view.
    \item The projection of the 3D hand silhouette should overlap with the 2D segmentation estimations in each view.
    \item The pose of each hand joint should conform to the kinematic constraints of the hand model.
\end{enumerate}
The variables subject to this optimization are the pose $\bm{\theta} \in \mathbb{R}^{16 \times 3}$ (rotations of 16 joints)
, the shape $\bm{\beta} \in \mathbb{R}^{10}$, and the root translation $\mathbf{R} \in \mathbb{R}^{3}$ of the parametric hand model: MANO \cite{romero2017embodied}.
MANO will deterministically map these three parameters to the 3D hand joints and mesh vertices, as:
\begin{equation}
    \mathbf{V}, \mathbf{J} = \mathcal{M}(\bm{\theta}, \bm{\beta}) +  \mathbf{R},
\end{equation}
where $\mathcal{M}$ is the MANO's linear skinning function.
Therefore, the optimization objective is formulated as:
\begin{equation}
    \min_{\bm{\theta}, \bm{\beta}, \mathbf{R}} \Big( \sum_{i=1}^{N} \big( \mathcal{L}_{\text{2D}}(\bm{\theta}, \bm{\beta}, \mathbf{R})_i + \mathcal{L}_{\text{seg}}(\bm{\theta}, \bm{\beta}, \mathbf{R})_i \big) + \mathcal{L}_{\text{kin}}(\bm{\theta}) \Big),
    \label{eq:optim}
\end{equation}
where $\mathcal{L}_{\text{2D}}$ is the mean squared error between the 2D keypoint estimations and the projected 3D joints,
$\mathcal{L}_{\text{seg}}$ is the binary cross entropy between the 2D segmentation estimations and the rendered 3D mesh silhouette,
and $\mathcal{L}_{\text{kin}}$ is the anatomical error of the hand model proposed by Yang \etal (Anatomical-Constrained MANO) \cite{yang2024cpf-tpami}.
We adopt a pre-trained network, CMR-SG \cite{chen2021cmr}, to predict 2D keypoints and 2D segmentation.
These two predictions serve as the pseudo ground-truth for optimization. We train CMR-SG on the DexYCB training set, and then use the network to make predictions on its test set.
We implement this optimization within the PyTorch framework. The optimization is conducted for 300 iterations per multi-view sample using Adam optimizer with an initial step size of 1e-2.
The step size decays by a factor of 0.5 when no improvement in the last 50 iterations.
We report the final results in \cref{table:quantitative}, row \texttt{5}, where the \POEMs model outperforms the optimization method in all metrics.
Moreover, the \model demonstrates \textbf{greater efficiency in terms of inference time}.
The average inference time for a single multi-view sample on one GPU is \textbf{0.07 seconds} (14.29 Hz) for the \POEMs, compared to \textbf{9.8 seconds} (total 300 iterations) for the optimization method.

\subsection{Ablation Study}\label{sec:ablation}
In this section, we conduct an ablation study to evaluate the effectiveness of the key components in the \model model.

\paraheading{Two-stage Design.}
Recalling the problem formulation in \cref{sec:formulation} - \cref{eq:formulation}, our model features a two-stage architecture. In the first stage, we estimate the root translation via heatmap estimation, followed by reconstructing the root-relative 3D hand mesh.
In contrast, one-stage models directly regress the absolute 3D hand vertices position in the world coordinate system. To evaluate the benefits of our two-stage design, we adapt the \POEMs model to incorporate a one-stage design, referred to as \textbf{\POEMs-1stage}.
Specifically, as we bind the world coordinate system to the first camera, we confine the camera's frustum space within a cuboid bounded by $[-1.2, 1.2]~ m$ in the first camera's $X$ (horizontal) axis, $[-1.2, 1.2] ~m$ in $Y$ (vertical) axis, and $[0.2, 2.0] ~m$ in $Z$ (depth) axis.
The hand vertices are directly regressed within these predefined boundaries.
To construct the basis points set $\mathbf{P}$, the standard \model usually determine its origin through root estimation  (see \cref{sec:root_estim}).
In \textbf{\POEMs-1stage}, without root estimation, we set this origin as the mean position of the hand's root from the training set.
After placing the basis points in the world, we normalize their coordinates within $[0, 1]$ for each axis based on the predefined boundaries.
To initialize the query points' position $\mathbf{X}$, we establish learnable initial positions ($\mathbb{R}^{799 \times 3}$) that are initialized by a uniform distribution within $[0, 1]$
and optimized alongside the network weights.
Thus, both the basis points and query points are represented in the normalized cuboid space. The remaining parts are identical with the original \POEMs model -- the Transformer decoder progressively refines the initial query points to form the final prediction.
Results on \dexycbmv dataset (see \cref{tab:ablation} row \texttt{2}) show large improvement with the standard \POEMs model over the one-stage approach.
Besides, since we have to manually decide the root position for each dataset, this \textbf{\POEMs-1stage} model would have less generalizability across different datasets and different camera-view configurations.

\paraheading{Point Cloud Embedding.}
This study focus on the improvement gains from point cloud embedding -- using a set of fixed points as the spatial basis for representing the query points in estimation.
In this variant of \POEMs model, we retain the first root estimation stage and replace the Point-Embedded Transformer with the Transformer decoder in the \textbf{PE-MeshTR}.
This ablation model is referred to as \textbf{\POEMs-w/o-pt}.
To illustrate the differences more clearly:
\begin{enumerate}[label=\roman*).,leftmargin=15pt]
    \item \textbf{Compared with \textbf{PE-MeshTR}}: \POEMs-w/o-pt includes an additional root estimation stage, meaning the query points are initialized and updated relative to the root position.
    \item \textbf{Compared with the \POEMs}:  \POEMs-w/o-pt omits the Projective Aggregation and Point-Embedded Transformer Decoder. Instead, the image features from multi-view are processed via Position Embedded Fusion (as in the \textbf{PE-MeshTR}), and served as the input embeddings for the Key ($K$) and Value ($V$) calculations in the DETR decoder.
\end{enumerate}
The results in \cref{tab:ablation} row \texttt{4,12} show that the Point Cloud Embedding leads superior performance.

\paraheading{\Basis's Projective Aggregation.}
This study evaluates the benefits derived from the Projective Aggregation of the basis points $\mathbf{P}$.
In this variant of the \POEMs model, each point in $\mathbf{P}$ can carry feature from only one image,
even though different points may carry features from different images.
To achieve this design, we first revisit the definition of the camera mesh grid, $\bm{M}_{cam}$ and the its aligned image feature, $\mathbf{B}_{cam}^{\wedge}$
from the Position Embedded Fusion (\cref{sec:main_exp}, PE-MeshTR).
Each cell in $\bm{M}_{cam}$ stores a 3D point coordinate, and the corresponding cell in $\mathbf{B}_{cam}^{\wedge}$
stores a feature vector of the same point.
After transforming $N$ camera mesh grids $\bm{M}_{cam}$ to the world space,
we can use the Ball Query to sample 4096 mesh grid points from the $N$ $\bm{M}_{\mathcal{W}}$.
Thanks to the randomness of Ball Query, these sampled points should originate from different camera's mesh grids -- \textit{different points carry features from different images}.
Simultaneously, each sampled point can only be derived from one camera (only corresponds to one image feature grid in
$\mathbf{B}_{cam}^{\wedge}$) -- \textit{each point carries features from only one image}.
We refer to this model as \textbf{\POEMs-w/o-proj}, indicating ``\textbf{without} \Basis's \textbf{proj}ective aggregation''.
Comparison between \cref{tab:ablation} row \texttt{3} \vs \texttt{1}, \texttt{9} \vs \texttt{6}, and \texttt{11} \vs \texttt{10} show the improvement brought by the Projective Aggregation.

\paraheading{\Query's Learnable Embeddings Selection}
In the standard DETR Transformer model, a set of learnable parameters, termed as Object Queries, is used to predict object classes and bounding boxes. This design is widely adopted in Transformer models for human and hand pose estimation, where a set of learnable parameters  (\eg query embeddings  $\mathbf{E_X}$), with lengths equivalent to the number of keypoints, predicts the positions of these keypoints. This study examines different strategies for constructing the query embeddings.
\begin{enumerate}[label=\roman*).,leftmargin=15pt]
    \item \textbf{V-emb} ($\mathbf{E^{\rm{V}}_X}$). Same as the Keypoint Transformer \cite{hampali2022kptrasnformer}, the $\mathbf{E^{\rm{V}}_X}$ is the vertex-level learnable embedding vectors, where $\mathbf{E^{\rm{V}}_X}\in \mathbb{R}^{799 \times d}$. This strategy corresponds to the default setting in the \POEMs model (see \cref{sec:learnable_query}).
    \item \textbf{G-emb} ($\mathbf{E^{\rm{G}}_X}$). Same as the Mesh Transformer \cite{lin2021metro}, the $\mathbf{E^{\rm{G}}_X}$ is the concatenation (\textcircled{c}) of a global image feature: $\mathbf{G} \in \mathbb{R}^{d}$ and a constant vertex-specified position: $\mathbf{X}_i^{\dagger}$ ($i$ for $i$-th vertex) extracted from a zero-posed and mean-shape MANO hand template ($\dagger$). Therefore, $\mathbf{E^{\rm{G}}_X}_i = \mathbf{G}~\textcircled{c}~\mathbf{X}_i^{\dagger}$, where $\mathbf{E^{\rm{G}}_X} \in \mathbb{R}^{799 \times (d+3)}$. Here, $\mathbf{G}$ is the average of the $N$ final-layer features from the $N$-views' image backbones.
    \item \textbf{G\&V-emb} ($\mathbf{E^{+}_X}$). Same as the MvP \cite{Wang2021MVP}, the $\mathbf{E^{+}_X}$ is the sum of global image feature $\mathbf{G}$, and vertex-level embedding vectors $\mathbf{E^{\rm{V}}_X}$, as $\mathbf{E^{+}_X}_{i} = \mathbf{G} + \mathbf{E^{\rm{V}}_X}_i$, where $\mathbf{E^{+}_X} \in \mathbb{R}^{799 \times d}$.
\end{enumerate}
The results in \cref{tab:ablation} row \texttt{6-8} show that the \textbf{V-emb} best fit for our model.

\subsection{\POEMg: Generalizable Multi-view HMR}\label{sec:scaling-up}

The \POEMs model, once trained, can only applies on certain camera configuration (including numbers, orders, and relative poses between cameras), significantly limiting its real-world application. To illustrate the shortcomings of \POEMs, we performed the following tests:
\begin{enumerate*}[label=\roman*)]
    \item Randomly shuffle the camera order in a test set (camera order change);
    \item Testing a model trained on one dataset using another dataset (camera numbers and relative poses change).
\end{enumerate*}
The results are shown in \cref{tab:robustness}
As observed, a model trained on \dexycbmv (8 cameras) performs notably worse when the camera order is shuffled. Similarly, a model trained on \hodddmv(5 cameras) performs poorly when tested on \oakinkmv (4 cameras) and vice versa.
To address these limitations of the \POEMs model, we introduce the \textbf{generalizable} multi-view hand mesh reconstruction model: \textbf{\POEMg}.
The virtue of \POEMg lies in scaling up data numbers for training, the data diversity for training, and the capacity of the deep network.

\begin{table}[!htp]
    \begin{center}
        \caption{Performance Degradation of \POEMs when changing the camera order, numbers, and poses.}\vspace{-3mm}
        \label{tab:robustness}
        \resizebox{1.0\linewidth}{!}{
            \begin{tabular}{llcccc}
                \toprule
                datasets                   & methods                                     & MPVPE$\downarrow$                       & PA$_{\rm{V}}$$\downarrow$               & MPJPE$\downarrow$                       & PA$_{\rm{J}}$$\downarrow$               \\
                \midrule
                \multirow{3}*{{DexYCB-Mv}} & \textbf{\POEMs}                             & \textbf{6.13}                           & \textbf{4.00}                           & \textbf{6.06}                           & \textbf{3.93}                           \\
                                           & cam. shuffle                                & 20.40                                   & 15.90                                   & 18.10                                   & 13.05                                   \\
                                           & \multicolumn{1}{r}{\footnotesize\% degrade} & \multicolumn{1}{r}{\footnotesize 233\%} & \multicolumn{1}{r}{\footnotesize 298\%} & \multicolumn{1}{r}{\footnotesize 199\%} & \multicolumn{1}{r}{\footnotesize 232\%} \\
                \noalign{\smallskip}
                \multirow{2}*{{OakInk-Mv}} & \textbf{\POEMs}                             & \textbf{6.20}                           & \textbf{4.21}                           & \textbf{6.01}                           & \textbf{4.00}                           \\
                                           & train.on HO3D                               & 29.9                                    & 20.5                                    & 28.6                                    & 18.3                                    \\
                                           & \multicolumn{1}{r}{\footnotesize\% degrade} & \multicolumn{1}{r}{\footnotesize 382\%} & \multicolumn{1}{r}{\footnotesize 387\%} & \multicolumn{1}{r}{\footnotesize 376\%} & \multicolumn{1}{r}{\footnotesize 358\%} \\
                \noalign{\smallskip}
                \multirow{2}*{{HO3D-Mv}}   & \textbf{\POEMs}                             & \textbf{17.20}                          & \textbf{9.97}                           & \textbf{17.28}                          & \textbf{9.60}                           \\
                                           & train.on OakInk                             & 30.60                                   & 13.9                                    & 30.40                                   & 12.85                                   \\
                                           & \multicolumn{1}{r}{\footnotesize\% degrade} & \multicolumn{1}{r}{\footnotesize 178\%} & \multicolumn{1}{r}{\footnotesize 139\%} & \multicolumn{1}{r}{\footnotesize 176\%} & \multicolumn{1}{r}{\footnotesize 134\%} \\
                \bottomrule
            \end{tabular}
        }
    \end{center}
\end{table}

\begin{figure*}[!htp]
    \centering
    \includegraphics[width=\linewidth]{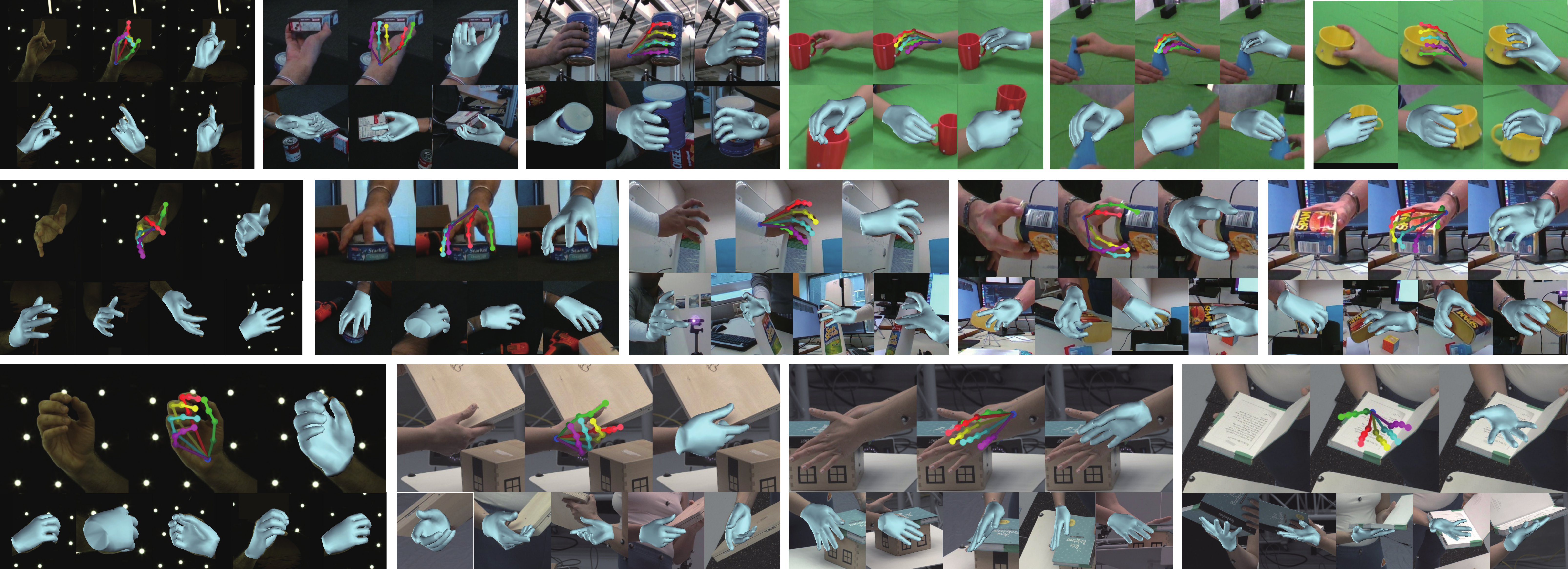}\vspace{-3mm}
    \caption{Qualitative results of the \POEMg model on the all five testing set. Each images patch contains a multi-view sample with the first row showing the primary camera view and the second row the rest.}
    \label{fig:poem_qualitative}
\end{figure*}

\begin{table*}[!t]
    \addtolength{\tabcolsep}{-1pt}
    \begin{center}
        \caption{Quantitative results of the Generalizable \POEMg model on all five multi-view testing set}\vspace{-3mm}
        \label{table:poemg_quantitative}
        \footnotesize
        \begin{tabular}{lllcccccccc}
            \toprule
            datasets                    & \#          & methods                                  & \MPVPE & \RRV  & \PAV & \AUCVtwenty & \MPJPE & \RRJ  & \PAJ & \AUCJtwenty \\
            \midrule
            \multirow{7}*{{DexYCB-Mv}}  & \texttt{1}  & \POEMg-small (8)                         & 7.48   & 8.84  & 4.46 & 0.63        & 7.22   & 8.78  & 4.16 & 0.63        \\
                                        & \texttt{2}  & \parbox{1.3cm}{\textbf{\POEMg}}(8 cams.) & 6.80   & 7.27  & 4.08 & 0.67        & 6.69   & 7.30  & 3.92 & 0.66        \\
                                        & \texttt{3}  & \parbox{1.3cm}{~}(4 cams.)               & 7.91   & 8.10  & 4.49 & 0.62        & 7.82   & 8.15  & 4.32 & 0.61        \\
                                        & \texttt{4}  & \parbox{1.3cm}{~}(2 cams.)               & 10.95  & 10.02 & 5.37 & 0.51        & 10.88  & 10.14 & 5.20 & 0.50        \\
                                        & \texttt{5}  & \parbox{1.3cm}{~}(rand. 2-8)             & 8.40   & 8.40  & 4.63 & 0.60        & 8.31   & 8.47  & 4.45 & 0.59        \\
            \noalign{\smallskip}
                                        & \texttt{6}  & \POEMg-large (8)                         & 6.50   & 7.50  & 3.84 & 0.68        & 6.33   & 7.54  & 3.68 & 0.68        \\
                                        & \texttt{7}  & \POEMg-param (8)                         & 7.00   & 7.46  & 4.26 & 0.66        & 6.95   & 7.59  & 4.21 & 0.65        \\
            \midrule
            \multirow{5}*{OakInk-Mv}    & \texttt{8}  & \POEMg-small (4)                         & 9.08   & 10.00 & 5.02 & 0.55        & 8.75   & 10.00 & 4.67 & 0.55        \\
                                        & \texttt{9}  & \parbox{1.3cm}{\textbf{\POEMg}}(4 cams.) & 8.55   & 8.58  & 4.68 & 0.58        & 8.34   & 8.60  & 4.42 & 0.58        \\
                                        & \texttt{10} & \parbox{1.3cm}{~}(rand. 2-4)             & 9.12   & 9.10  & 4.89 & 0.56        & 8.92   & 9.13  & 4.62 & 0.55        \\
            \noalign{\smallskip}
                                        & \texttt{11} & \POEMg-large (4)                         & 8.33   & 8.47  & 4.49 & 0.59        & 8.09   & 8.47  & 4.24 & 0.59        \\
                                        & \texttt{12} & \POEMg-param (4)                         & 7.91   & 7.51  & 4.08 & 0.61        & 7.75   & 7.65  & 3.95 & 0.60        \\

            \midrule
            \multirow{5}*{HO3D-Mv}      & \texttt{13} & \POEMg-small (5)                         & 8.53   & 12.23 & 5.36 & 0.58        & 7.97   & 12.27 & 4.82 & 0.59        \\
                                        & \texttt{14} & \parbox{1.3cm}{\textbf{\POEMg}}(5 cams.) & 8.12   & 10.83 & 5.16 & 0.60        & 7.72   & 10.91 & 4.72 & 0.60        \\
                                        & \texttt{15} & \parbox{1.3cm}{~}(rand. 2-5)             & 9.58   & 11.70 & 5.57 & 0.55        & 9.22   & 11.78 & 5.15 & 0.55        \\
            \noalign{\smallskip}
                                        & \texttt{16} & \POEMg-large (5)                         & 7.63   & 10.58 & 4.59 & 0.63        & 7.18   & 10.58 & 4.13 & 0.63        \\
                                        & \texttt{17} & \POEMg-param (5)                         & 9.16   & 11.27 & 5.72 & 0.55        & 8.89   & 11.38 & 5.39 & 0.55        \\

            \midrule
            \multirow{3}*{InterHand-Mv} & \texttt{18} & \POEMg-small (8)                         & 9.69   & 11.91 & 6.86 & 0.60        & 9.12   & 11.73 & 6.22 & 0.58        \\
                                        & \texttt{19} & \parbox{1.3cm}{\textbf{\POEMg}}(8 cams.) & 9.45   & 11.37 & 6.58 & 0.62        & 9.05   & 11.35 & 6.18 & 0.59        \\
                                        & \texttt{20} & \POEMg-large (8)                         & 7.64   & 9.04  & 5.23 & 0.69        & 7.19   & 8.82  & 4.72 & 0.67        \\
            \midrule
            \multirow{3}*{ARCTIC-Mv}    & \texttt{21} & \POEMg-small (8)                         & 7.09   & 9.53  & 5.50 & 0.62        & 7.67   & 9.20  & 4.81 & 0.62        \\
                                        & \texttt{22} & \parbox{1.3cm}{\textbf{\POEMg}}(8 cams.) & 6.78   & 8.08  & 5.13 & 0.67        & 6.34   & 7.80  & 4.63 & 0.67        \\
                                        & \texttt{23} & \POEMg-large (8)                         & 5.89   & 7.13  & 4.43 & 0.71        & 5.42   & 6.80  & 3.91 & 0.71        \\
            \bottomrule
        \end{tabular}
    \end{center}
\end{table*}

In the first experiment of this section,
we compare the performance of the \POEMg model on the \dexycbmv, \hodddmv, and \oakinkmv test sets with three separately-trained \POEMs models.
The results indicate that \POEMg exhibits the most significant improvement on \hodddmv.
This can be attributed to the small training sample size in \hodddmv,
where \POEMg leverages the generalization capability of a larger model and larger dataset
to better adapt to this camera configuration. Additionally,
we observe that \POEMg does not perform as well on \oakinkmv, which aligns with our expectations -- systematic biases exist in any data annotation method, and models trained solely on the \oakinkmv are better at fitting these biases. Since \POEMg involves diverse datasets (each with different annotation biases), its ability to overfit a specific datasets decreases. We also present the performance of \POEMg on the \arcticmv, \interhandmv test sets in \cref{table:poemg_quantitative} row \texttt{18-23}. \POEMg demonstrates consistent performance across all six test sets.

In the second experiment, we reduce the number of cameras in the test sets to observe the performance change.  We gradually decrease the number of cameras from the total (up to 8) to 2, with several cameras being randomly removed, and the remaining cameras' order re-shuffled. The results, shown in \cref{table:poemg_quantitative} row \texttt{2-5}, \texttt{9-10}, and \texttt{14-15} indicate that \POEMg performs better with more cameras. Even though, significantly reducing the number of cameras does not catastrophically affect its performance, thereby overcoming the limitations of \POEMs.

In the third experiment, we explore the relationship between model size and \POEMg's performance.
We designed three model sizes by varying the dimension of token features in the Transformer:
\POEMg-small ($d$=128), \POEMg-base ($d$=256, default \POEMg), and \POEMg-large ($d$=512).
Results show that performance improves with increasing model size.

The final experiment introduces a variant of \POEMg: the parametric POEM model, \textbf{\POEMg-param}. In this variant, we replace the output of the Transformer decoder in the second stage of \POEMg with the MANO model parameters $\bm{\theta}, \bm{\beta}$ instead of directly outputting hand vertex positions. The parametric output directly provides rotation angles for each joint, making it more suitable for VR/AR applications that require hand animation. The results are presented in \cref{table:poemg_quantitative} row \texttt{7,12,17}.

\subsection{Comparisons with Single-View Methods under Procrustes Alignment}
To provide a holistic benchmark, we also compare our method with state‑of‑the‑art single‑view 3D hand reconstruction frameworks.
Since the absolute position of hands is ambiguous in a single-view setting, we only report the MPJPE and MPVPE under the Procrustes Alignment,which removes absolute translation, wrist rotation, and aligns the hand scale.

We perform this comparison on the \textbf{official HO3D test set} (\textbf{v2} \cite{hampali2020ho3dv2} and \textbf{v3} \cite{hampali2021ho3dv3}) and benchmark against a suite of single‑view methods that reconstruct a 3D hand mesh either by regressing parametric model coefficients or by directly predicting mesh vertices.
The model we report is \POEMg-large.
Notably, this model is trained on a mixture of six datasets (refer to \cref{sec:datasets}); its HO3D component comes from our \textbf{HO3D‑Mv} training sequences, while evaluation uses the official HO3D v2 \& v3 testing sequences (see \cref{tab:ho3d_eval_set}).
\vspace{-0.7em}
\begin{table}[!h]
    \newcolumntype{L}[1]{>{\raggedright\arraybackslash}p{#1}}
    \centering
    \caption{The HO3Dv2\&v3 sequence IDs for \POEMg training and evaluations.}\vspace{-3mm}
    \label{tab:ho3d_eval_set}
    \footnotesize
    \begin{tabular}{L{70pt} L{60pt} L{70pt}}
        \toprule
        train                        & HO3D-Mv eval & official HO3D eval  \\
        \midrule
        ABF1, BB1, GSF1, MDF1, SiBF1 & GPMF1, SB1   & AP1, MPM1, SB1, SM1 \\
        \bottomrule
    \end{tabular}
\end{table}

Following the official HO3D challenge protocol, we also report the AUC (@50mm) of joint and vertex error and F-scores (@5mm and @15mm) of vertex error. The comparison results are presented in \cref{tab:ho3d_eval} and \cref{tab:ho3dv3_eval}

\begin{table}[!t]
    \addtolength{\tabcolsep}{-3pt}
\centering
\caption{Comparison with single-view SOTAs on HO3Dv2 official benchmark}\vspace{-3mm}
\label{tab:ho3d_eval}%
\footnotesize
\begin{tabular}{@{}lcccccc@{}}
\toprule
method & \PAJ & \AUCJ & \PAV & \AUCV & F@5$\uparrow$ & F@15$\uparrow$ \\ 
\midrule
Pose2Mesh~\cite{choi2020pose2mesh}        & 12.5 & 0.754 & 12.7 & 0.749 & 0.441 & 0.909 \\
ArtiBoost~\cite{yang2022artiboost}        & 11.4 & 0.773 & 10.9 & 0.782 & 0.488 & 0.944 \\
I2L-MeshNet~\cite{moon2020i2l}            & 11.2 & 0.775 & 13.9 & 0.722 & 0.409 & 0.932 \\
Hasson~\etal~\cite{hasson2019learning}    & 11.0 & 0.780 & 11.2 & 0.777 & 0.464 & 0.939 \\
Keypoint Trans~\cite{hampali2022kptrasnformer} & 10.8 & 0.786 & –    & –     & –     & –     \\
Hampali~\etal~\cite{hampali2020ho3dv2}& 10.7 & 0.788 & 10.6 & 0.790 & 0.506 & 0.942 \\
METRO~\cite{lin2021metro}                   & 10.4 & 0.792 & 11.1 & 0.779 & 0.484 & 0.946 \\
Liu~\etal~\cite{liu2021semihand}              & 9.9  & 0.803 & 9.5  & 0.810 & 0.528 & 0.956 \\
I2UV-HandNet~\cite{chen2021i2uv}          & 9.9  & 0.804 & 10.1 & 0.799 & 0.500 & 0.943 \\
MobRecon~\cite{chen2022mobrecon}          & 9.2  & –     & 9.4  & –     & 0.538 & 0.957 \\
HandOccNet~\cite{Park2022HandOccNet}      & 9.1  & 0.819 & 8.8  & 0.819 & 0.564 & 0.963 \\
AMVUR~\cite{jiang2023probabilistic}      & 8.3  & 0.835 & 8.2  & 0.836 & 0.608 & 0.965 \\
HandBooster~\cite{xu2024handbooster} & 8.2 & 0.836 & 8.4 & 0.832 & 0.585 & 0.972 \\
HaMeR~\cite{Pavlakos2024hamer}  & 7.7  & 0.846 & 7.9  & 0.841 & 0.635 & 0.980 \\ 
Hamba~\cite{dong2024hamba} & 7.5 & 0.850 & 7.7 & 0.846 & 0.648 & 0.982 \\
\midrule
\textbf{\POEMg-large} & \textbf{6.3} & \textbf{0.874} & \textbf{7.2} & \textbf{0.856} & \textbf{0.666} & \textbf{0.989} \\
\bottomrule
\end{tabular}
\end{table}%

\begin{table}[!t]
    \addtolength{\tabcolsep}{-3pt}
\centering
\caption{Comparison with single-view SOTAs on HO3Dv3 official benchmark}\vspace{-3mm}
\label{tab:ho3dv3_eval}%
\footnotesize
\begin{tabular}{@{}lcccccc@{}}
\toprule
method & \PAJ & \AUCJ & \PAV & \AUCV & F@5$\uparrow$ & F@15$\uparrow$ \\ 
\midrule
S$^2$HAND~\cite{chen2021s2hand}       & 11.5 & 0.769 & 11.1 & 0.778 & 0.448 & 0.932 \\
Keypoint Trans~\cite{hampali2022kptrasnformer}       & 10.9 & 0.785 & -    & -     & -     & -     \\
ArtiBoost~\cite{yang2022artiboost}         & 10.8 & 0.785 & 10.4 & 0.792 & 0.507 & 0.946 \\
AMVUR~\cite{jiang2023probabilistic}            & 8.7  & 0.826 & 8.3  & 0.834 & 0.593 & 0.964 \\
Hamba~\cite{dong2024hamba}    & 6.9 & 0.861 & 6.8 & 0.864 & 0.681 & 0.982 \\
HandOS~\cite{chen2025handos}  & 6.8 & - & 6.7 & - & 0.688  & 0.983 \\
\midrule
\textbf{\POEMg-large} & \textbf{5.8} & \textbf{0.884} & \textbf{6.3} & \textbf{0.874} & \textbf{0.715} & \textbf{0.989} \\
\bottomrule
\end{tabular}
\end{table}%

\subsection{Real-world Application}\label{sec:application}
\begin{figure}[!b]
    \centering
    \includegraphics[width=1.0\linewidth]{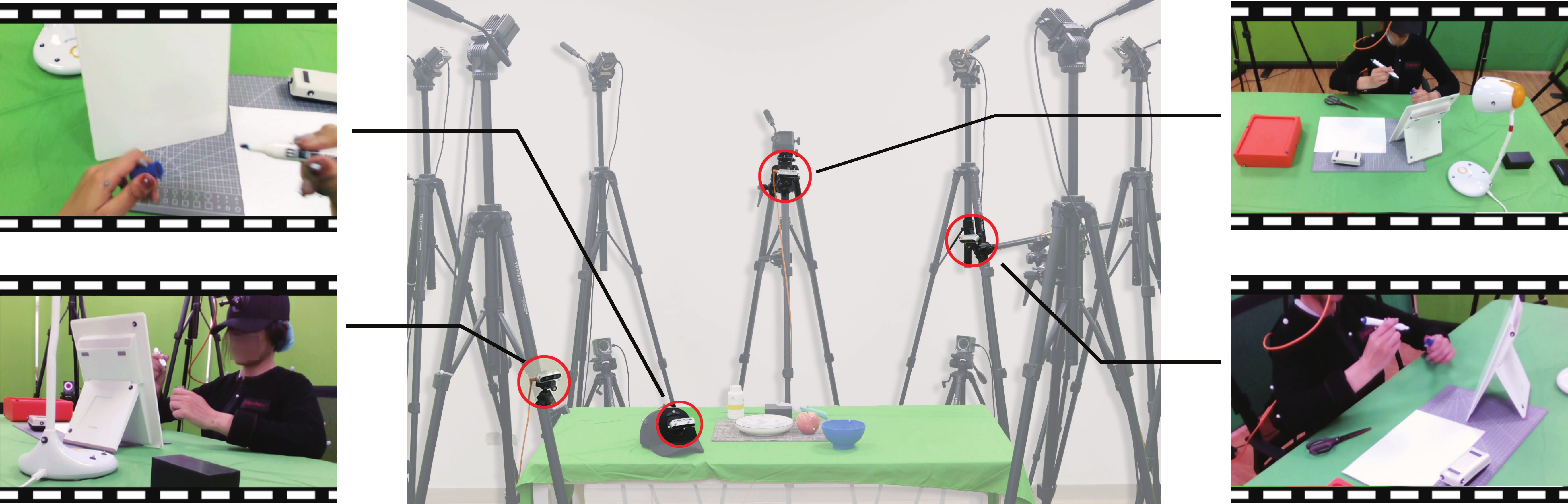}\vspace{-3mm}
    \caption{The multi-view hand tracking platform used in real-world application. The four RGB cameras circled in red capture the users bi-manual hand motion, while the other 12 MoCap cameras are used for obtaining the ground-truth.}
    \label{fig:capture_setup}
\end{figure}
In this section, we deploy the pre-trained \POEMg model in a real-world application to examine its potential for practical use. The aim of this application is to accurately reconstruct the user's bi-manual hands motion (sequence of poses) during a real-world manipulation task.
We aim to achieve this goal \textbf{using only the \POEMg model and several user-adjustable RGB cameras}.

\begin{figure}[!htp]
    \centering
    \includegraphics[width=\linewidth]{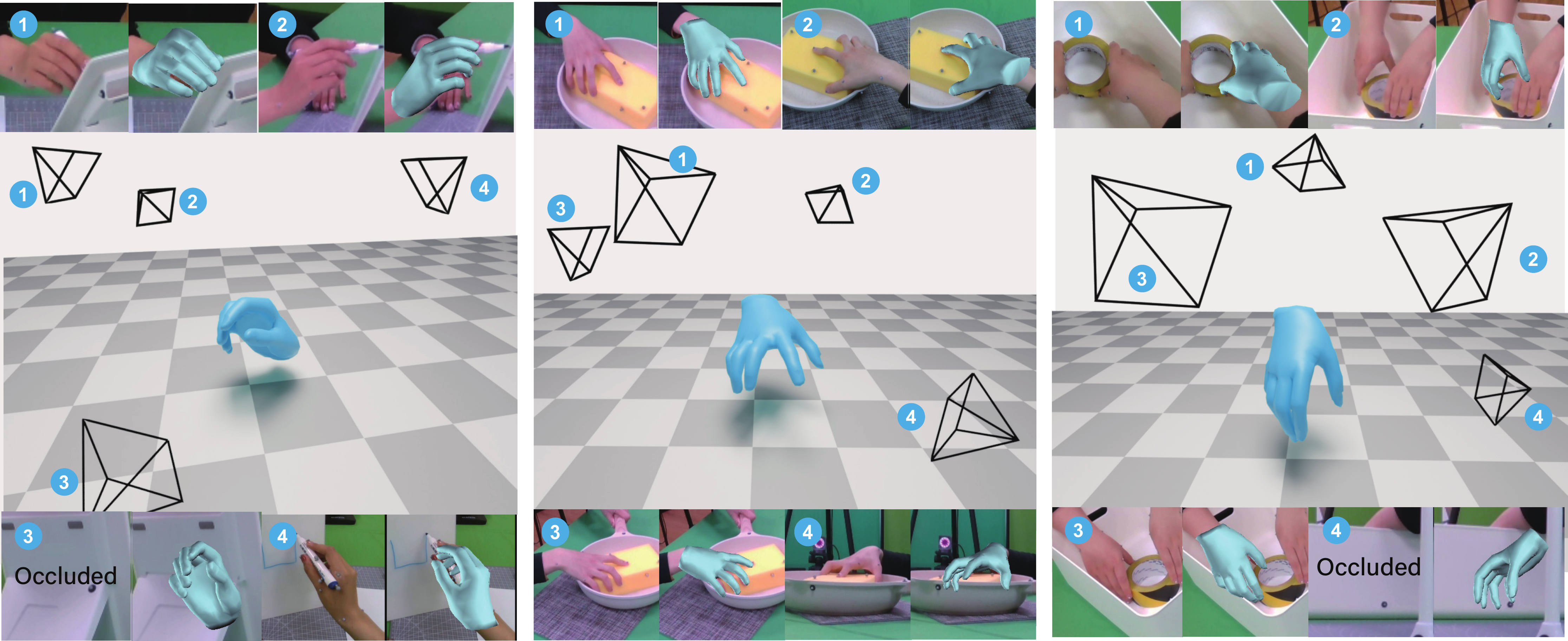}\vspace{-3mm}
    \caption{Real-world application: the camera configuration and the estimated hand mesh in 3D space and each view.}
    \label{fig:app_3d_layout}
\end{figure}

\begin{figure*}[!htp]
    \centering
    \includegraphics[width=\linewidth]{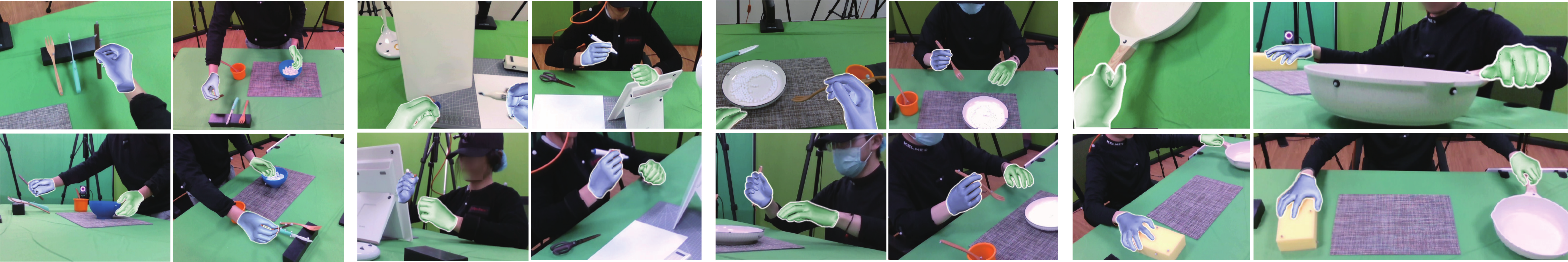}\vspace{-3mm}
    \caption{Real-world application: POEM supports mesh reconstruction of \textbf{both the left and right hands}. The results are shown as overlaid hand mesh renderings on each view.}
    \label{fig:app_both_hands}
\end{figure*}

To this end, we construct an experimental setup, as shown in \cref{fig:capture_setup}. The setup consists of four RGB cameras: three positioned for third-view observations and one mounted on the user's hat for a first-view observation. To obtain ground-truth for the user's hand poses, we construct a MoCap system consisting of 12 OptiTrack infrared cameras, and attach 11 reflective hemispheres on the skin of the user's hands. We choose the hemispheres with diameters of 4 mm to minimize their interfere with the user's hand movements and reduce their visibility in the RGB cameras. By tracking the reflective hemispheres, the MoCap system provides the 3D positions of these points.
We employ the Mosh++ pipeline \cite{mahmood2019amass} to fit the MANO model to these 3D points, yielding positions for 778 hand vertices. These vertices provide the ground-truth for evaluating the performance of \POEMg on real-world data.
For the camera extrinsic calibration, we also attach reflective hemispheres to the outer casing of the four RGB cameras. This enables the MoCap system to track the transformations of these cameras relative to the world.

During the recording of the test set, users are instructed to perform a series of bi-manual tasks on the experimental table, such as writing with a pen and opening a bottle. The multi-view observations of the right hand are directly input into the same \POEMg model used in \cref{sec:scaling-up}. For the left hand, we apply a world-mirroring process to the multi-view observations. Specifically, this involves horizontally flipping all images and mirroring camera poses along the world-camera (first camera)'s Y-Z plane. The mirrored observations are then fed into the \POEMg model as right hand. After the model outputs the 3D positions of 799 query points, we mirror back the 3D points along the Y-Z plane to obtain the final 3D positions for the left hand.

\begin{table}[!tp]
    \addtolength{\tabcolsep}{-2pt}
    \begin{center}
        \caption{Comparison between \POEMg and 2D keypoint-based  optimization on the in-the-wild test set.}\vspace{-3mm}
        \label{tab:realworld}
        \footnotesize
        \begin{tabular}{lcccc}
            \toprule
            methods                          & \MPVPE         & \PAV          & \MPJPE         & \PAJ          \\
            \midrule
            {\footnotesize MediaPipe-optim.} & 16.26          & 7.72          & 16.51          & 8.08          \\
            \textbf{\POEMg}-base             & \textbf{12.57} & \textbf{6.71} & \textbf{12.08} & \textbf{6.30} \\
            \bottomrule
        \end{tabular}
    \end{center}
\end{table}

In addition to the \POEMg model, we utilize another off-the-shelf foundation perception model: MediaPipe Hand Tracking \cite{zhang2020mediapipe}, for comparison. MediaPipe is also widely used for obtaining hand keypoints during data collection \cite{tian2024gaze, qin2023anyteleop}. We first use MediaPipe’s pre-trained palm detection and hand landmark detection models to predict the 2D positions of 21 hand joints in the images. For images with valid 2D predictions, we utilize the iterative optimization method mentioned in \cref{sec:experiments} to fit the MANO parameters, and ultimately obtain the positions of the 3D joints and mesh vertices.
As shown in \cref{tab:realworld}, the performance of the \POEMg model notably outperforms traditional solutions represented by MediaPipe.

Finally, we present some qualitative results from the \POEMg model, as illustrated in \cref{fig:app_3d_layout,fig:app_both_hands}. It can be observed that the \POEMg model performs commendably well on in-the-wild data, excelling in both 3D global positioning and 2D image alignment, and is capable of reconstruct both left and right hand simultaneously.


\section{Discussion}\label{sec:discussion}
To successfully deploy \model model into real-world scenarios, several prerequisites must be satisfied. Firstly, a detection module is necessary to localize the hand within images and differentiate between the left and right hands. Secondly, the extrinsic and intrinsic camera parameters must be provided.

When these conditions are met, \model integrates seamlessly into various real-world multi-camera setups. Typical scenarios include: 1) VR headsets equipped with multi-camera arrays, and 2) dome-like dexterous robotic teleoperation systems.

Provided that  sufficient disparity in fields-of-view (FoV) exists among different cameras, the model can performs as intended.
Specifically, FoV disparity is crucial for calculating the absolute root position (as described in \cref{eq:dlt}). After determining the root point, the model constructs basis points around it, followed by projective aggregation to sample features from multiple views.
During deployment, our method does not require the cameras to be arranged in a spherical distribution; it only requires that the spherical set of basis points is observable (\ie projected onto image planes) from the available cameras.

Once the absolute root translation is known -- even from a \textbf{single view} -- our model can still \textbf{reasonably reconstruct} the relative positions of all 778 hand vertices with respect to the root.
This robustness stems from the design of POEM model’s architecture: referring to \cref{eq:proj_aggr}, we \textbf{simulate a single-view shortcut} during training. This path bypasses cross-view feature aggregation and allows features from an individual view to be directly propagated through the network. This design forces the model to learn useful reconstruction cues from each individual view, rather than depending entirely on cross-view fusion. To validate this, we select one sequence (SM1) from the official HO3Dv2 test set, where root translation is provided by the dataset's ground truth. In this sequence, our \POEMg model achieves a PA-MPJPE of 7.4 mm and a PA-MPVPE of 7.9 mm.
As shown in \cref{tab:ho3d_eval}, these results do not degrade significantly compared to the full-view setup.
As expected, the model performs better when more effective views are available (refer to rows \texttt{2-5} in \cref{table:poemg_quantitative}).

In several common extreme multi-view scenarios, the robustness of our model can be expected:
\begin{enumerate*}[label=\textbf{\roman*}).,leftmargin=15pt]
    \item When a hand is visible in one camera’s FoV but significantly occluded by objects, our model’s attention mechanism in multi-view feature aggregation effectively compensates by utilizing visible information from other cameras. Consequently, the prediction is robust to partial occlusions in individual viewpoints (as illustrated by the occluded-view predictions in \cref{fig:app_3d_layout}).
    \item If the hand is entirely outside the observable range of some cameras (with detection returning empty), since our model naturally handles inputs of varying camera counts, it discards invisible viewpoints during the forward pass, ensuring a reasonable output.
\end{enumerate*}

Future work proceeds along two avenues.
\begin{enumerate*}[label=\textbf{\roman*}).,leftmargin=15pt]
    \item We aim to eliminate the requirements on hand detector and external camera calibration, achieving truly one‑stage multi‑view HMR.
    \item We plan to jointly regress the hand mesh and its contact points, enabling contact‑rich motion capture.
\end{enumerate*}

\ifCLASSOPTIONcompsoc
\section*{Acknowledgments}
This work was supported by the National Key Research and Development Project of China (Grant 2022ZD0160102 and 2021ZD0110704), STCSM Science and Technology Innovation Action Plan (Grant 23511103104 and 24511103202), STCSM "Qi Ming Xing - Yang Fan" Talent Program (Grant 24YF2722000), Science and Technology Major Project of Jiangsu Province (Grant BG2024041), and Shanghai Artificial Intelligence Laboratory, XPLORER PRIZE grants.
\else
\section*{Acknowledgment}
\fi





%






{
  \bibliographystyle{IEEEtran}
  \bibliography{egbib}
}

\vspace{-6mm}
\begin{IEEEbiography}
    [{\includegraphics[width=1in,height=1.25in,clip,keepaspectratio]{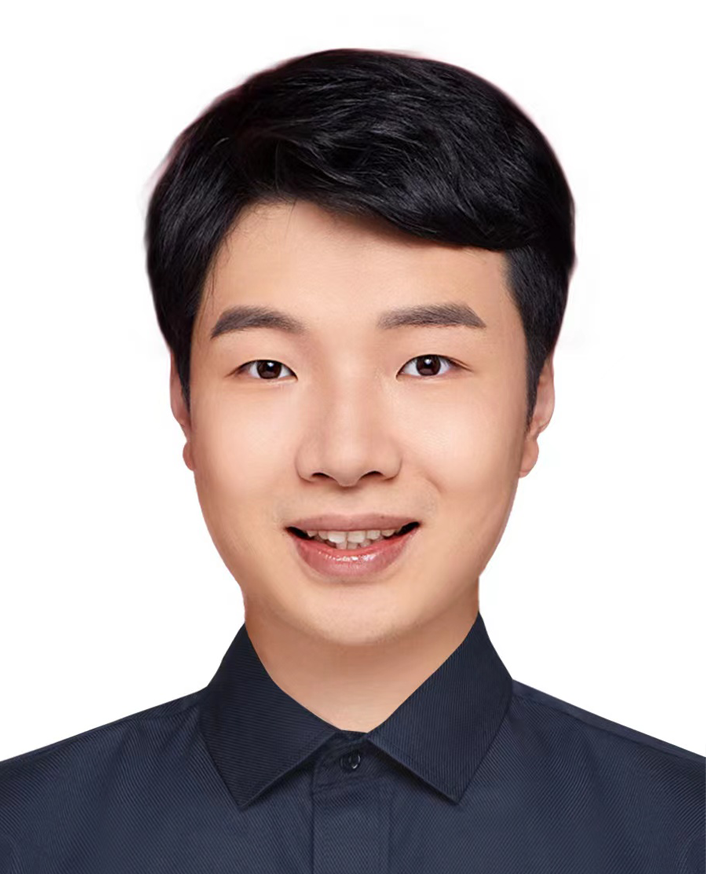}}]{Lixin Yang} is an Assistant Research Professor at Shanghai Jiao Tong University (SJTU). He
    got his Ph.D. from Machine Vision and Intelligence Group, SJTU in 2023, supervised by Prof. Cewu Lu, and his M.S. degree from the Intelligent Robot Lab, SJTU in 2019. His research interests include 3D computer vision and embodied AI. More information: \url{https://lixiny.github.io}
\end{IEEEbiography}

\vspace{-5mm}
\begin{IEEEbiography}
    [{\includegraphics[width=1in,clip,keepaspectratio]{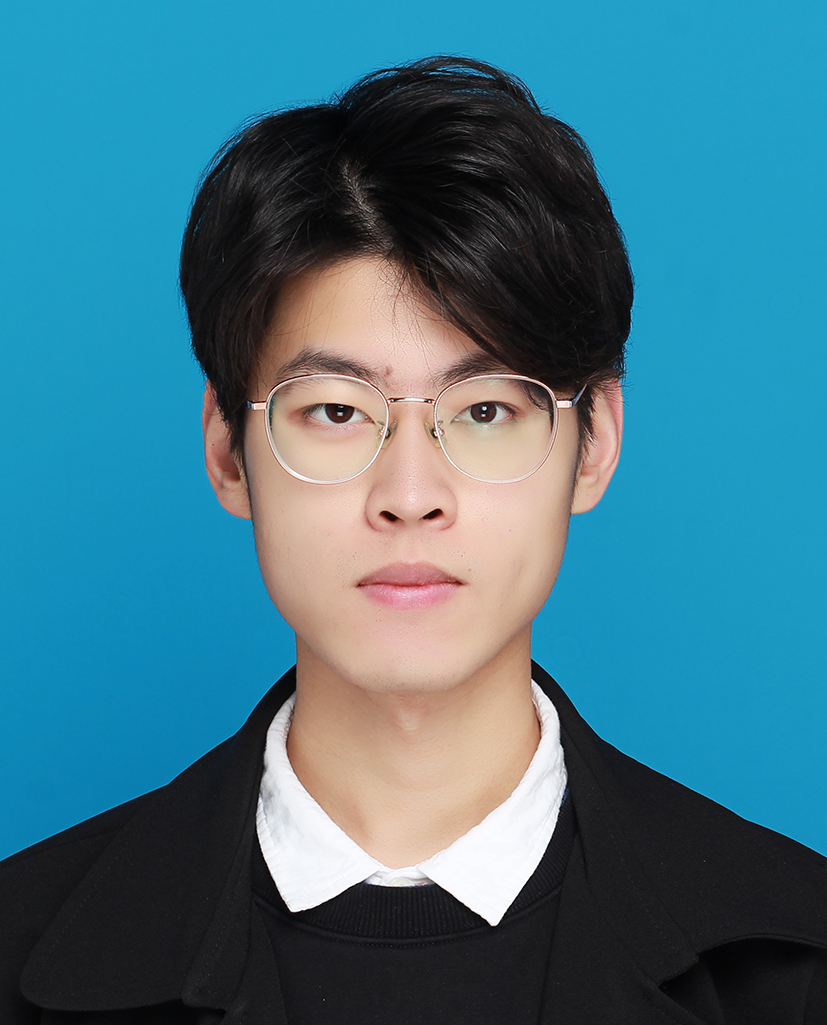}}]{Licheng Zhong} is a Research Assistant at Shanghai Qi Zhi Institute. He got his bachelor’s degree at School of Mechanical Engineering, Shanghai Jiao Tong University in 2024. His research interests include 3D Vision, AIGC and Embodied AI. More information: \url{https://zlicheng.com}
\end{IEEEbiography}

\vspace{-5mm}
\begin{IEEEbiography}
    [{\includegraphics[width=1in,clip,keepaspectratio]{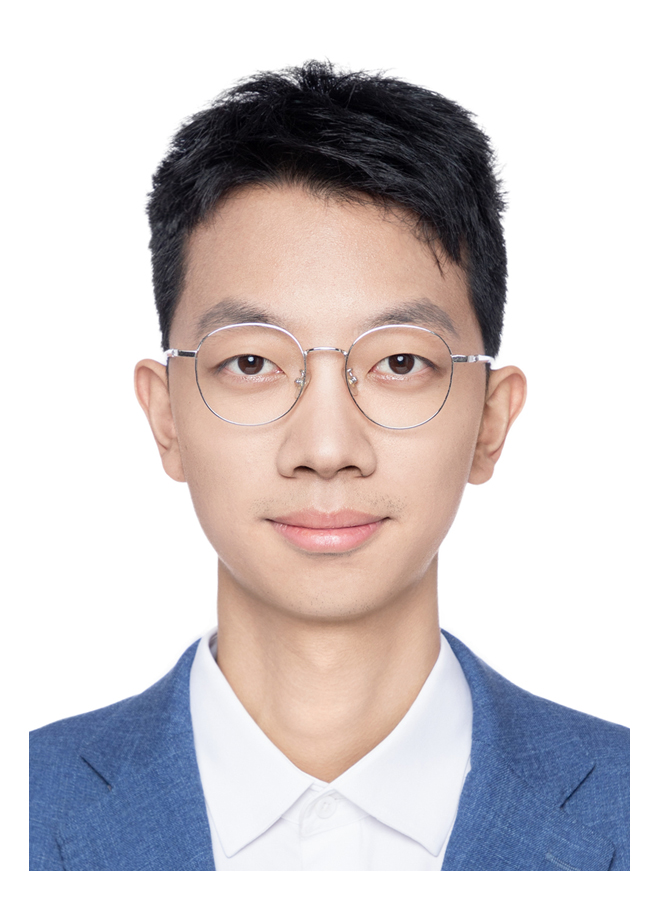}}]{Pengxiang Zhu} is an undergraduate student at Shanghai Jiao Tong University (SJTU). He is currently pursuing the bachelor’s degree at Department of Computer Science in SJTU. His research interests include 3D computer vision and embodied intelligence. He has been a member of Machine Vision and Intelligence Group (MVIG), supervised by Prof. Cewu Lu and Lixin Yang. More information available at \url{https://jubsteven.github.io}
\end{IEEEbiography}

\vspace{-5mm}
\begin{IEEEbiography}
    [{\includegraphics[width=1in,height=1.25in,clip,keepaspectratio]{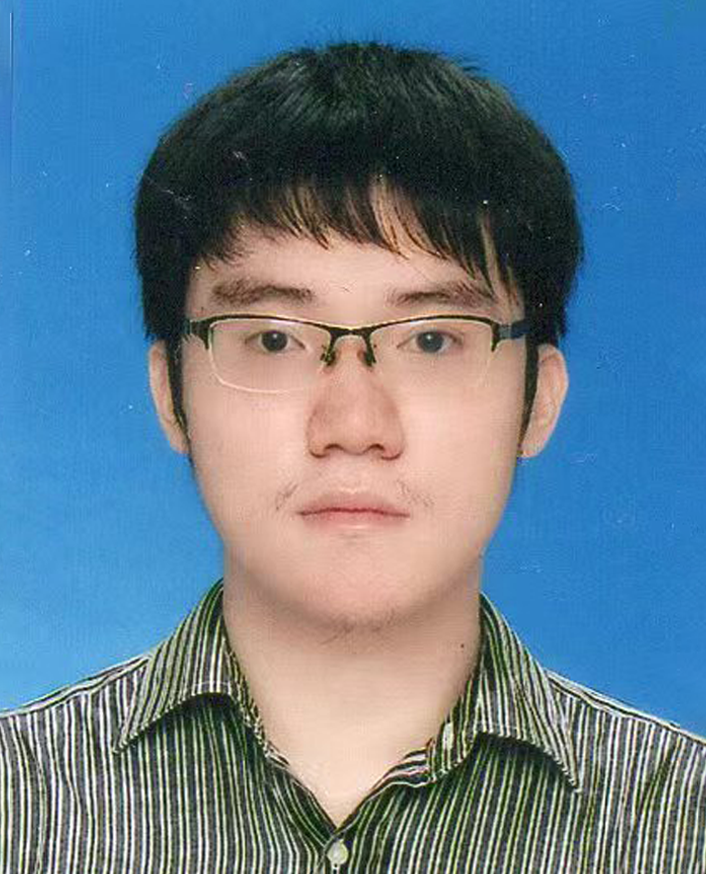}}]{Xinyu Zhan} is a Ph.D student at Shanghai Jiao Tong University, Shanghai, China. His research interests include 3D computer vision and embodied AI. Zhan received his bachelor degree in computer science from Shanghai Jiao Tong University. He is a member of MVIG-SJTU.
\end{IEEEbiography}

\vspace{-5mm}
\begin{IEEEbiography}
    [{\includegraphics[width=1in,height=1.25in,clip,keepaspectratio]{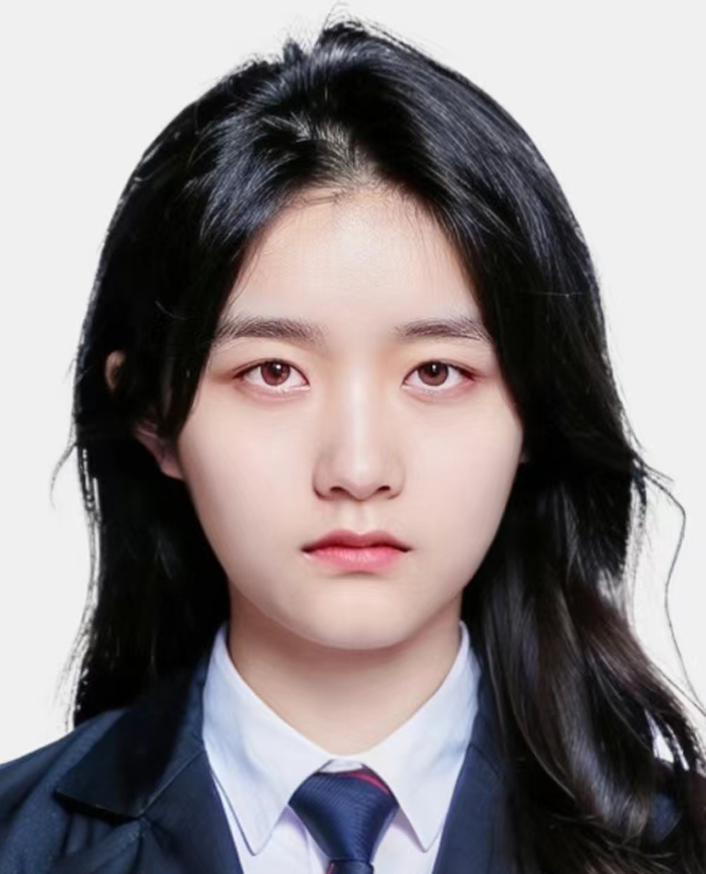}}]
    {Junxiao Kong} is an undergraduate at Shanghai Jiao Tong University (SJTU), studying at the School of Electronic Information and Electrical Engineering (SEIEE). Her research interests lie in Embodied AI and 3D Computer Vision. She is a member of the Machine Vision and Intelligence Group (MVIG), under the mentorship of Prof. Cewu Lu and Lixin Yang.
\end{IEEEbiography}

\vspace{-5mm}
\begin{IEEEbiography}
    [{\includegraphics[width=1in,height=1.25in,clip,keepaspectratio]{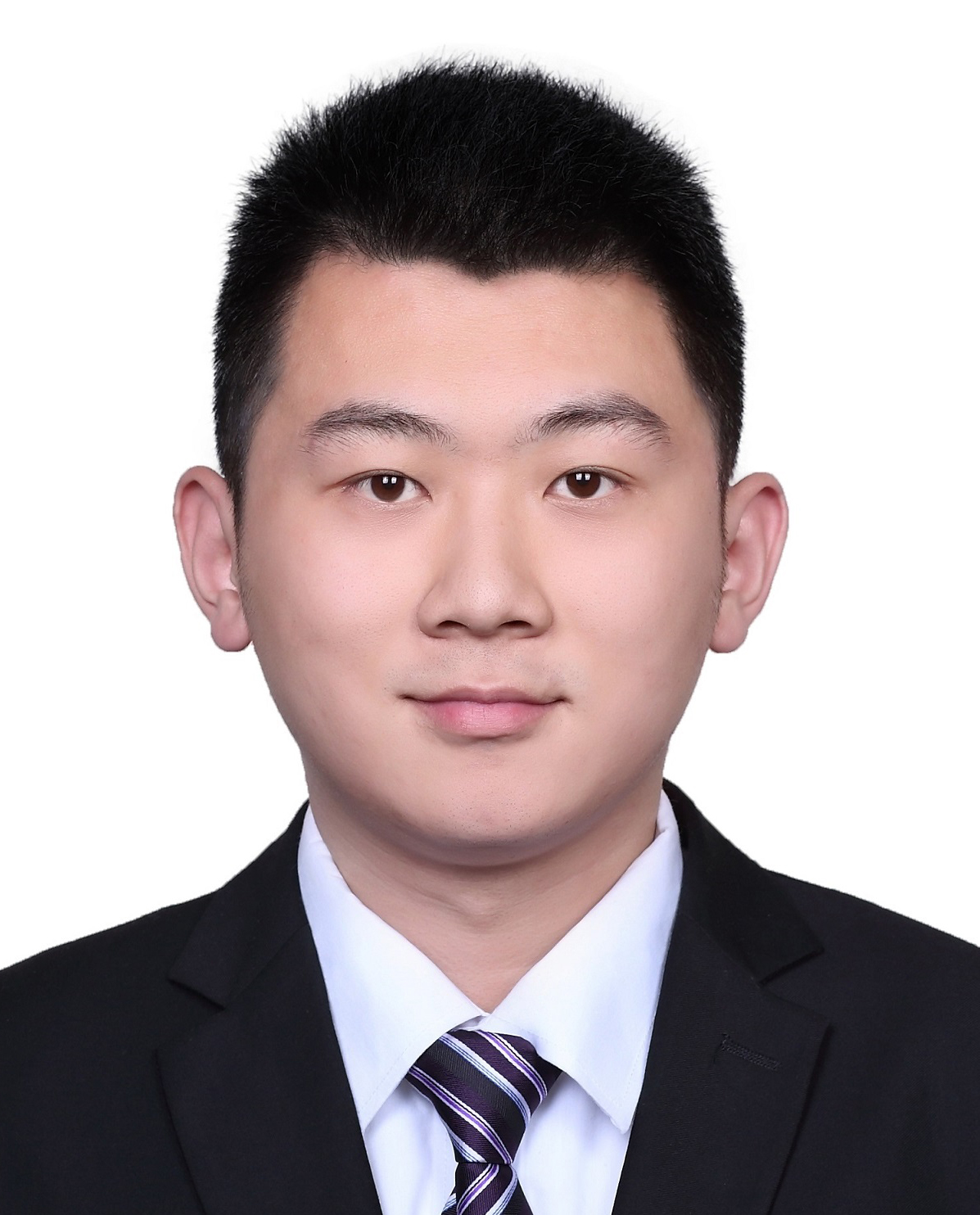}}]
    {Jian Xu} is an Associate Professor at Institute of Automation Chinese Academy of Sciences (CASIA). His research interests include pose estimation and large multimodal models. He obtained his Ph.D. degree from CASIA.
\end{IEEEbiography}

\vspace{-5mm}
\begin{IEEEbiography}
    [{\includegraphics[width=1in,height=1.25in,clip,keepaspectratio]{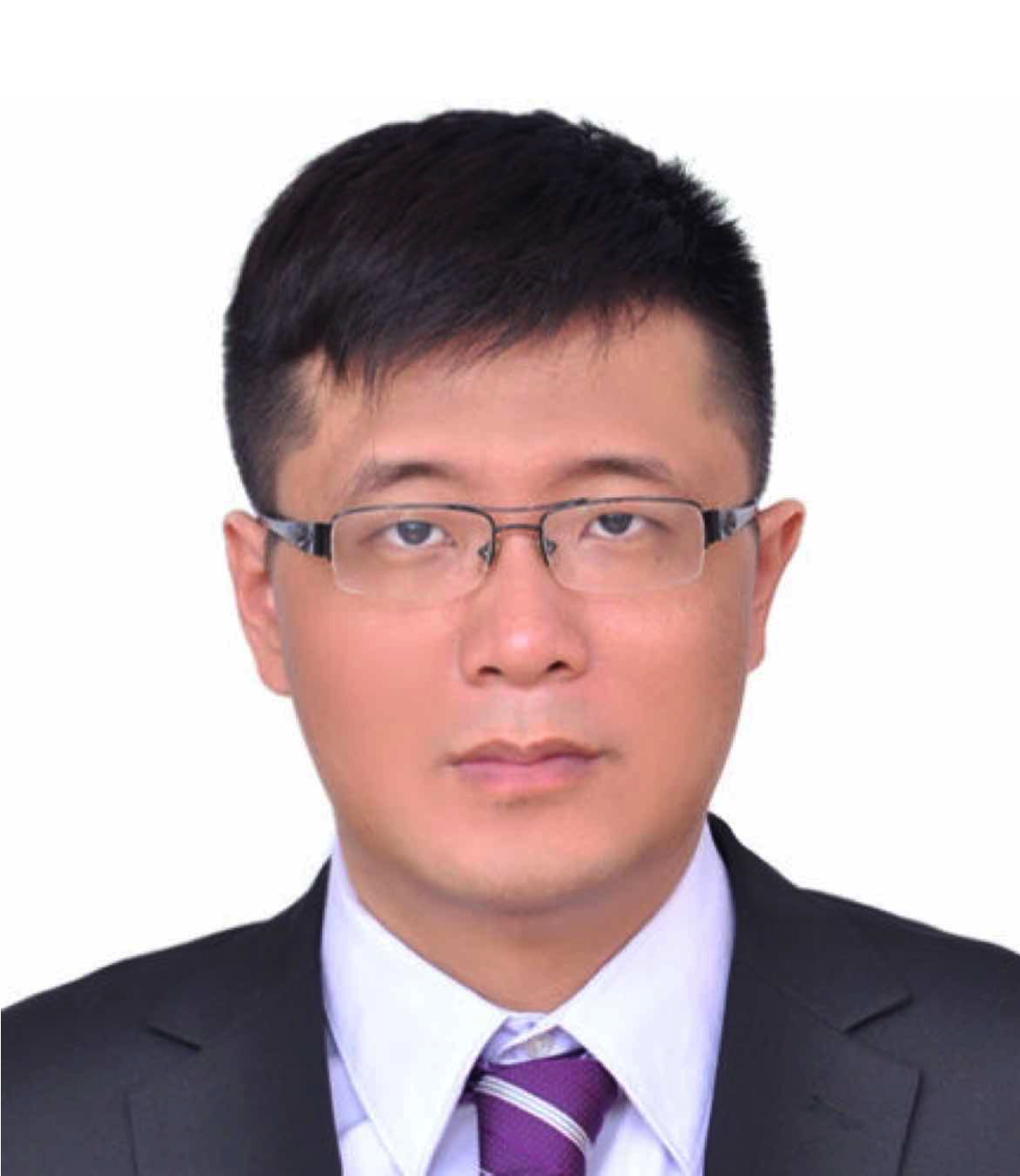}}]{Cewu Lu} is a Professor at Shanghai Jiao Tong University (SJTU). Before he joined SJTU, he was a research fellow at Stanford University, working under Prof. Fei-Fei Li and Prof. Leonidas J. Guibas. He was a Research Assistant Professor at Hong Kong University of Science and Technology with Prof. Chi Keung Tang. He got his Ph.D. degree from the Chinese University of Hong Kong, supervised by Prof. Jiaya Jia. He is one of the core technique members in Stanford-Toyota autonomous car project.
    He serves as an associate editor for journal CVPR and reviewer for journal TPAMI and IJCV. His research interests fall mainly in computer vision, deep learning, deep reinforcement learning and robotics vision.
\end{IEEEbiography}

%



\end{document}